\documentclass[journal]{IEEEtran}
\ifCLASSINFOpdf
\else
\fi
\usepackage{tikz}
\usepackage{makecell}
\usepackage{multirow}
\usepackage{hhline}
\usepackage{pgfplots}
\usetikzlibrary{patterns}
\usetikzlibrary{matrix}
\usepackage{enumitem}
\usepackage{amsmath, amsthm, amssymb}

\newcommand {\highlight}[1]{\textcolor[rgb]{0,0,0} {#1}}
\newcommand {\mycomment}[1]{\textcolor[rgb]{1,0,0} {#1}}
\newcommand{\tabincell}[2]{\begin{tabular}{@{}#1@{}}#2\end{tabular}} 
\newcolumntype{L}[1]{>{\raggedright\arraybackslash}p{#1}}
\newcolumntype{C}[1]{>{\centering\arraybackslash}p{#1}}
\newcolumntype{R}[1]{>{\raggedleft\arraybackslash}p{#1}}
\newcommand {\minorhighlight}[1]{\textcolor[rgb]{0,0,0} {#1}}

\usepackage{comment}

\begin{document}
%
\title{Attended End-to-end Architecture for Age Estimation from Facial Expression Videos}
%
%
%

\author{Wenjie~Pei,
        Hamdi~Dibeklio\u{g}lu, \emph{Member, IEEE},
        Tadas~Baltru\v{s}aitis,
        and~David~M.J.~Tax
\thanks{W. Pei is with the Harbin Institute of Technology, Shenzhen, China (e-mail: wenjiecoder@gmail.com and homepage: https://wenjiepei.github.io/).} 
\thanks{H. Dibeklio\u{g}lu is with the Department of Computer Engineering, Bilkent University, Ankara 06800, Turkey (e-mail: dibeklioglu@cs.bilkent.edu.tr).}
\thanks{T. Baltru\v{s}aitis is with Microsoft Corporation, Microsoft Research Lab Cambridge, United Kingdom CB1 2FD (e-mail: tabaltru@microsoft.com)}
\thanks{D. Tax is with the Pattern Recognition Laboratory, Delft University of Technology, Delft 2628 CD, the Netherlands (e-mail: D.M.J.Tax@tudelft.nl).} 
}

\maketitle

\begin{abstract}
The main challenges of age estimation from  facial expression videos lie not only in the modeling of the static facial appearance, but also in the capturing of the temporal facial dynamics. Traditional techniques to this problem focus on constructing handcrafted features to explore the discriminative information contained in facial appearance and dynamics separately. This relies on sophisticated feature-refinement and framework-design. In this paper, we present an end-to-end architecture for age estimation, called Spatially-Indexed Attention Model (SIAM), which is able to simultaneously learn both the appearance and dynamics of age from raw videos of facial expressions. Specifically, we employ convolutional neural networks to extract effective latent appearance representations and feed them into recurrent networks to model the temporal dynamics. More importantly, we propose to leverage attention models for salience detection in both the spatial domain for each single image and the temporal domain for the whole video as well. We design a specific spatially-indexed attention mechanism among the convolutional layers to extract the salient facial regions in each individual image, and a temporal attention layer to assign attention weights to each frame. This two-pronged approach not only improves the performance by allowing the model to focus on informative frames and facial areas, but it also offers an interpretable correspondence between the spatial facial regions as well as temporal frames, and the task of age estimation. We demonstrate the strong performance of our model in experiments on a large, gender-balanced database with 400 subjects with ages spanning from 8 to 76 years. Experiments reveal that our model exhibits significant superiority over the state-of-the-art methods given sufficient training data.
\end{abstract}


%
\IEEEpeerreviewmaketitle

\section{Introduction}

\IEEEPARstart{H}{uman} age estimation from faces is an important research topic due to its extensive applications ranging from surveillance monitoring~\cite{GuoFDH08,Lanitis2004} to forensic art~\cite{forensic,fu2010age} and social networks~\cite{gallagher09,gallagher08}. 
The widely-used discriminative features for age estimation are appearance-related, such as wrinkles in the face, skin texture and luster, hence plenty of prevalent methods focus on modeling the appearance information from the static face~\cite{Han2013,Eirikur17}. Recent studies~\cite{DibekliogluICM2012, Hamdi2015} also indicate that the dynamic information in facial expressions like temporal properties of a smile can be leveraged to significantly improve the performance of age estimation. It is reasonable since there are intuitive temporal dynamics involved in facial expressions which are relevant to the age. For instance, the exhibited facial movement like wrinkles in the smiling process is different for people of different ages.    

The traditional approaches to age estimation from facial expression videos focus on constructing handcrafted features to explore the discriminative information contained in static appearance and temporal dynamics separately, and then combining them into an integrated system~\cite{DibekliogluICM2012, Hamdi2015}. However, these kinds of methods rely on sophisticated feature 
design. In this work, we propose a novel end-to-end architecture for age estimation from facial expression videos, which is able to automatically learn the static appearance in each single image and the temporal dynamics contained in the facial expression simultaneously. In particular, we employ convolutional neural networks (CNNs) to model the static appearance due to its successful representation learning in the image domain. The learned latent appearance features for each image are subsequently fed into recurrent networks to model the temporal dynamics. In this way, both static appearance and temporal dynamics can be integrated seamlessly in an end-to-end manner. A key benefit of this design is that the learned static appearance features and temporal dynamic features are optimized jointly,
which can lead to better performance for the final task of age estimation. Additionally, the end-to-end manner of our method avoids separating feature design from the age regression as a two-stage procedure, which is the typical way of the methods using handcrafted features.  

Attention models have been proposed to let models learn by themselves to pay attention to specific regions in an image~\cite{Kelvin2015} or different segments in a sequence~\cite{pei2017temporal,Bahdanau2015} according to the relevance to the aimed task. Likewise, different facial parts (in each single image) and different phases of the expression (in the inspected video) may exhibit varying degrees of salience to age estimation. 
To incorporate attention, we customize a specific attention module for spatial facial salience detection. To detect temporal salience in the sequential expression, we mount a temporal attention module upon the recurrent networks. It serves as a filtering layer to determine the amount of information of each frame to be incorporated into final age regression task. 
All functional modules of our proposed Spatially-Indexed Attention Model (SIAM) including convolutional networks for learning appearance, recurrent networks for learning dynamics, two attention modules as well as the final age regression module can be trained jointly, without any manual intervention. 

Extensive experiments on a real-world database demonstrate the substantial superiority of our model over the state-of-the-art methods. Our model has the capacity to learn and it could do even better on more data while other models potentially saturate and do not get better no matter how much data you give them.
Notably, larger training data tends to explore more potential of our model and expand its advantages compared to other methods.

\section{Related Work}
In this study, we propose to effectively learn spatial and temporal patterns of aging in an attended end-to-end manner for a more reliable age estimation. To comprehend the related concepts, in this section, the literature on automatic age estimation will be summarized, and an overview of neural attention models will be given.

\begin{figure*}[!tb]
 \begin{center}
 \includegraphics[width=0.95\linewidth]{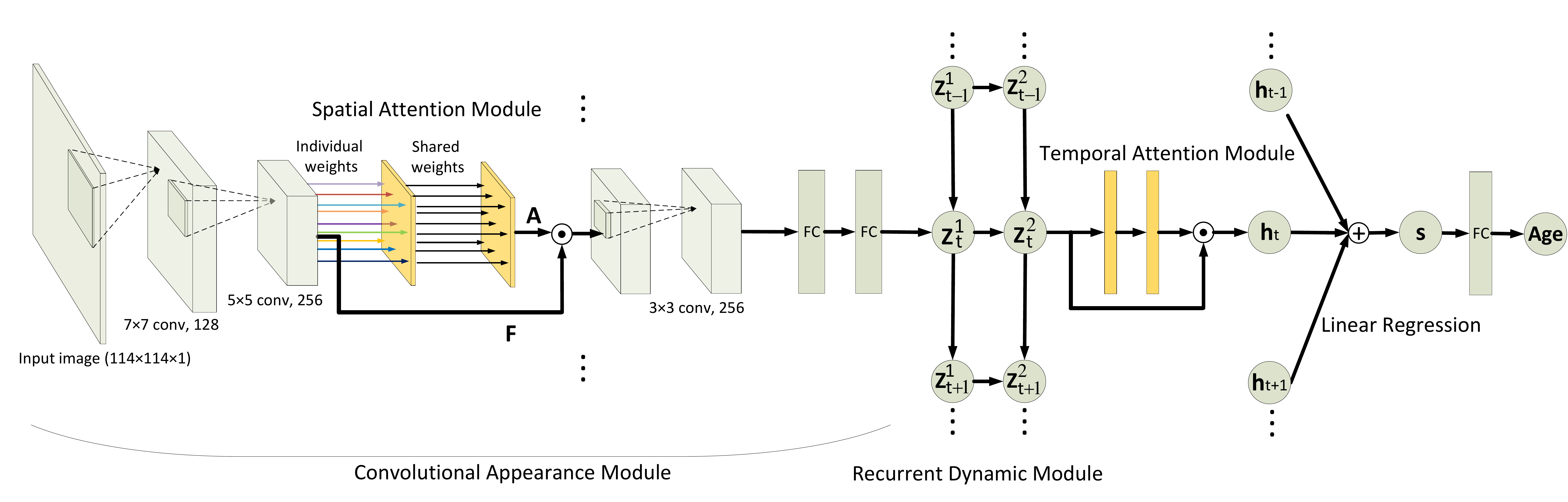} 
 \end{center}
 \caption{The graphical representation of our Spatially-Indexed Attention Model (SIAM) for age estimation. Note that we only give one instance of the convolutional appearance module, the spatial attention module and temporal attention module for one frame at time step $t$. Other frames share exactly the same structure for all three modules.}
 \label{fig:model}
\end{figure*}

\subsection{Automatic Age Estimation}
Based on the fact that age information is crucial to understand requirements or preferences of people, automatic estimation of age from face images has quite a few real-life applications, and thus, it has been a very active area of research over the last two decades. Biometric search/authentication in facial image databases, denying purchase of unhealthful products (e.g. alcohol and tobacco) by underage customers, and personalization/adaptation of interactive systems to the users (displaying personalized advertisements) can be counted as some examples of the use of age estimation.

One of the major requirements in facial age estimation is the capturing/modeling of facial properties that change by age. These patterns are related to craniofacial development~\cite{alley2013social} and alteration in facial skin (e.g. wrinkles)~\cite{lanitis2002toward}. While the majority of earlier studies in the area focus on describing such patterns using engineered (handcrafted) representations such as local binary patterns (LBP)~\cite{Ojala_LBP} and its variations~\cite{ylioinas2013age}, Gabor filter based biologically-inspired aging features (BIF)~\cite{guo2009human}, and shape-based features~\cite{thukral2012hierarchical}, some studies propose to capture aging patterns through learning-based approaches such as subspace learning~\cite{Guo08, Gen06, Gen08, zhan2011age, chen2013subspace}, PCA-tree-based encoding~\cite{alnajar2012learning}, and metric learning~\cite{chao2013facial, li2015ordinal}. 

The increase in the number and size of facial age databases, and the recent dramatic improvements in the field of deep learning have shifted the focus towards deep architectures to learn complex (nonlinear) patterns of facial aging from a large collection of face images. For instance, \cite{wang2015deeply} presents the first exploration of employing CNNs for age estimation, where representations obtained from different layers of CNN are used. In \cite{levi2015age}, a multi-task CNN architecture is proposed to optimize the facial representation jointly for age and gender estimation. \cite{gurpinar2016kernel} models the features extracted from a pre-trained CNN (VGG-Face~\cite{parkhi2015deep}) using the kernel extreme learning machines. \cite{rothe2016deep} uses VGG-16 CNN architecture~\cite{simonyan2014very} (pre-trained on ImageNet images) and fine-tunes the model using a multi-class classifier for age estimation. Then, softmax-normalized output probabilities are used for the final prediction. Differently from conventional methods, \cite{rothe2016deep} solely employs face detection for face alignment rather than using facial landmark detection, leading to a more accurate age estimation. In \cite{agustsson2017anchored}, Agustsson~\textit{et al.} complement \cite{rothe2016deep} with their proposed Anchored Regression Network (rather than employing softmax classifier on top the VGG-16 CNN architecture), enhancing the reliability. In a recent study~\cite{xing2017diagnosing}, Xing~\textit{et al.} have analyzed different loss functions and CNN architectures for age estimation as well as employing a joint optimization together with race and gender classification tasks. 
\minorhighlight{Unlike previous methods that use deep neural networks in a supervised manner, \cite{wang2015age} investigates an unsupervised learning framework for CNN features by applying k-means to learn convolutional filters of a single-layer CNN. Obtained CNN features are refined by subsequent unsupervised recurrent neural networks (with randomly initialized parameters) and projected into a discriminative subspace for age estimation by training an SVM or SVR.}

In contrast to regression and multi-class classification, some studies approach age estimation as an ordinal ranking problem. For instance, \cite{BMVC2015} presents a deep (category-based) ranking model that combines deep scattering transform and ordinal ranking. \cite{niu2016ordinal} formulates the problem as ordinal regression using a series of binary classification tasks which are jointly optimized by a multiple output CNN architecture. In \cite{chen2017using}, instead of a multiple output model, a series of basic CNNs are employed (a separate CNN for each ordinal age group), and their binary outputs are aggregated. Such an approach allows capturing different patterns for different age groups.

The use of deep architectures has significantly improved the reliability of automatic age estimation, especially under pose and illumination variations. Facial expression variation, however, is still a challenge since expressions form deformations on the facial surface that can be confused with aging-related wrinkles. Yet, only a few recent works in the literature explore solutions for this issue~\cite{guo2012study, zhang2013age, guo2013facial, lou2017expression}. Guo \textit{et al.}~\cite{guo2012study} model correlation between aging features of the neutral face and a specific expression (i.e. smile) of individuals. Learned correlation is used to map the features of expressive faces to those of neutral ones. In this way, the confusing influence of expressions in aging patterns are removed. However, this method requires an accurate estimation of facial expressions (before the age estimation), and a separate training for each expression of interest using neutral and expressive face images of each subject in the training dataset. \cite{zhang2013age} learns a common subspace for a set of facial expressions that reduce the influence of expressions while preserving the aging patterns. \cite{guo2013facial} defines four age groups, and models each facial expression in each age group as a different class for cross-expression age estimation. In a similar manner, \cite{lou2017expression} proposes a multi-task framework that jointly learns the age and expression using the latent structured support vector machines. 

Interestingly, until a recent work of Hadid~\cite{hadid2011analyzing}, none of the methods in the literature have used facial videos for age estimation. In~\cite{hadid2011analyzing}, the volume local binary patterns (VLBP) features are employed to describe spatio-temporal information in videos of talking faces for age grouping (child, youth, adult, middle-age, and elderly). Yet, this video-based method (\cite{hadid2011analyzing}) could not perform as accurate as the image-based methods. On the other method, more recently, Dibeklio\u{g}lu \textit{et al.} have presented the first successful example of video-based age estimation~\cite{DibekliogluICM2012}, where displacement characteristics of facial landmarks are represented by a set of handcrafted descriptors extracted from smile videos, and combined with spatio-temporal appearance features. In a follow-up study, Dibeklio\u{g}lu \textit{et al.}~\cite{Hamdi2015} have enhanced their engineered features so as to capture temporal changes in 3D surface deformations, leading to a more reliable age estimation. It is important to note that these two studies exploit the aging characteristics hidden in temporal dynamics of facial expressions, rather than reducing the influence of expressions in aging features. Thus, following the informativeness of temporal dynamics of facial expressions and based on the success of deep models, the current study proposes a deep temporal architecture for automatic age estimation. 

Because the literature on automatic age estimation is extensive, for further information, we refer the reader to \cite{ramanathan2009computational}, \cite{fu2010age}, and to the more recent \cite{panis2016overview}.
\subsection{Attention models}
Much of progress in neural networks was enabled by so called neural attention, which allows the network to focus on certain elements of a sequence \cite{pei2017temporal,Bahdanau2015,Yao2015} or certain regions of an image \cite{Kelvin2015} when performing a prediction. The appeal of such models comes from their end-to-end nature, allowing the network to learn how to attend to or align data before making a prediction. 

It has been particularly popular in encoder-decoder frameworks, where it was first introduced to better translate between languages~\cite{Bahdanau2015}. The attention network learned to focus on particular words or phrases when translating sentences, showing large performance gains on especially long sequences. It has also been extensively used for visual image and video captioning, allowing the decoder module to focus on parts of the image it was describing \cite{Kelvin2015}. Similarly, the neural attention models have been used in visual question answering tasks, helping the alignment between words in the question and regions in the image~\cite{YangHGDS16}. Spatial Transformer Networks which focus on a particular area of image can also be seen as a special case of attention \cite{Jaderberg2015}. Somewhat relatedly, work in facial expression analysis has explored using particular regions for facial action unit detection \cite{Li2017, Zhao2016}, however, they did not explore dynamically attending to regions depending on the current facial appearance. Our work is inspired by these attention models, but explores different ways of constructing neural attention and applying it to age estimation.

\section{Method}

Given a video displaying the facial expression of a subject, the aim is to estimate the age of that person. Next to that, the model is expected to capture the salient facial regions in the spatial domain and the salient phase during facial expression in the temporal domain.  
Our proposed Spatially-Indexed Attention Model (SIAM) is composed of four functional modules: 1) a convolutional appearance module for appearance modeling, 2) a spatial attention module for spatial (facial) salience detection, 3) a recurrent dynamic module for facial dynamics, and 4) a temporal attention module for discriminating temporal salient frames. The proposed model is illustrated in Figure~\ref{fig:model}. We will elaborate on the four modules in a bottom-up fashion and explain step by step how they are integrated into an end-to-end trainable system. 

\subsection{Convolutional Appearance Module}
Convolutional neural networks (CNNs) have achieved great success for automatic latent representation learning in the image domain. We propose to employ CNNs to model the static appearance for each image of the given video. Compared to conventional handcrafted image features which are generally designed independently of the aimed task, the features learned automatically by CNNs tend to describe the aimed task more accurately due to the parameters learning by back-propagation from the loss function of the aimed task.

Our model contains three convolutional layers with coarse-to-fine filters and subsequent two fully-connected layers. The output of the last fully-connected layer is fed as input to recurrent modules at the corresponding frame. Response-normalization layers~\cite{Krizhevsky2012} and Max-pooling layers follow the first and second convolutional layers. The ReLU~\cite{ReLU} function is used as the nonlinear activation function for each convolutional layer as well as the fully-connected layer. We found that creating deeper networks did not lead to a performance improvement, possibly due to comparatively small training dataset used. It should be noted that the spatial attention module presented subsequently in Section~\ref{sec:spatial_att} is embedded among convolutional layers to perform facial salience detection. The details are depicted in Figure~\ref{fig:model}.

For each frame, the input image with size $114 \times 114 \times 1$ is filtered by the first convolutional layer with $128$ kernels with size $7 \times 7$ and stride size $2$. The second convolutional layer filters the output of the first convolutional layer with $256$ kernels of size $5 \times 5$. The spatial attention module (Section~\ref{sec:spatial_att}) then takes as input the output of the second convolutional layer and extracts the salient regions in the face. The generated attended feature map containing salience is filtered by the third convolutional layer that has $256$ kernels of size $3 \times 3$. The first fully-connected layer has $4096$ neurons while the second fully-connected layer has the same number of neurons as the subsequent recurrent network (as will be explained in Section~\ref{sec:temp_rnn}). 

It should be noted that the same convolutional module is shared across all frames in the time domain. Thus, the forward pass of the convolutional module can be computed in parallel for all frames. In the backward pass, the parameters in the convolutional module are optimized by back-propagating output gradients of the upper recurrent module through all frames.  

\subsection{Spatial Attention Module}
\label{sec:spatial_att}
The goal of the spatial attention module is to dynamically estimate the salience and relevance of different image portions for the downstream task (in our case age estimation). 
It is implemented as a feature map filter embedded after one of the convolutional layers in the convolutional appearance modules to preserve the information based on the calculated saliency score. 

Formally, suppose the output volume $\mathbf{F}$ of a convolutional layer $L$ has dimensions $M \times N \times C$, with $C$ feature maps of size $M \times N$. The spatial attention module embedded after the convolutional layer $L$ is denoted by a matrix $\mathbf{A}$ with the same size $M \times N$ as the feature map. The element $\mathbf{A}_{ij}$ of $\mathbf{A}$ indicates the attention weight (interpreted as saliency score) for the feature vector $\mathbf{F}_{ij}$ composed of $C$ channels located at $(i, j)$ (i.e., $|\mathbf{F}_{ij}| \equiv C$) in the feature map. Each feature vector corresponds to a certain part of the input image (i.e., receptive field). Therefore, the receptive field of the attention becomes larger when the attention module is inserted in latter convolutional layers. Section~\ref{sec:effect_att_position} presents an experimental comparison of the different positions of spatial attention module in the convolutional appearance module. In practice, we insert the spatial attention module after the second convolutional layer. 

We propose a spatially-indexed attention mechanism to model $\mathbf{A}$. Concretely, the attention weight $\mathbf{A}$ is modeled by two fully-connected layers: the first layer is parameterized by individual weights for each entry of feature map while the second layer shares the transformation weights across the whole map. Thus the attention weight $\mathbf{A}_{ij}$ is modeled as:
\begin{equation}
\mathbf{A}_{ij} = \sigma( \mathbf{u}^\top \tanh(\mathbf{W}_{ij} \mathbf{F}_{ij}+\mathbf{b}_{ij})+c)
\label{eqn:spatial_att1}
\end{equation}
Herein, $\mathbf{W}_{ij} \in \mathbb{R}^{d \times C}$ is the transformation matrix for the first fully-connected layer and $\mathbf{u}^\top \in \mathbb{R}^{d}$ is the weight vector for the second fully-connected layer to fuse the information from different channels and $c$ is a bias term. A sigmoid function $\sigma$ is employed as the activation function at the top layer of the attention module to constraint the attention weight to lie in the interval $[0, 1]$.  
The obtained attention matrix $\mathbf{A}$ controls the information flowing into the subsequent layer in the convolutional appearance module by an element-wise multiplication to each channel (feature map) of F:
\begin{equation}
\mathbf{I} = \mathbf{F} \odot \mathbf{A}
\label{eqn:spatial_att2}
\end{equation}
Here $\mathbf{I}$ is the output of the spatial attention module, which is fed into the subsequent layer of the convolutional appearance module.

It is worth mentioning that we use individual weights ($\mathbf{W}_{ij}$) for the first layer and shared weights $\mathbf{u}$ for the second fusion layer in the spatial attention module (this will be called a spatially-indexed mechanism). The first layer is expected to capture the local detailed (fine-grained) variation while the second fusion layer is designed to capture the global variation and smooth the attention distribution. It is different from the design of the soft attention model for image caption generation~\cite{Kelvin2015}, in which the transformation weights are shared in both two layers of the attention model. In that scenario, the attention model is used to capture the related objects in the input image to each word of generated caption and the objects are always easily separable from the background scene in the image. By contrast, we aim to capture the salient parts in a face image, which requires to model more detailed variation. Employing shared weights in both layers tends to blur the spatial variation. Besides, the typical attention model is translation-invariant. Namely, if the picture is rearranged, the attention would be very similar, whereas our attention is spatially-indexed. Section~\ref{sec:spatial_attention_mechanism} provides an comparison between different attention mechanisms by visualizing the learned attention weight distribution.       

\subsection{Recurrent Dynamic Module}
\label{sec:temp_rnn}
The temporal facial dynamics are expected to contribute to age estimation, which has been demonstrated by Dibeklio{\u g}lu et al.~\cite{Hamdi2015}.
In contrast to the handcrafted dynamics features they use~\cite{Hamdi2015}, we propose to employ recurrent networks to capture the underlying temporal information automatically. The potential advantages of using recurrent networks are that (1) they learn relevant dynamics feature to the aimed task (age estimation) smoothly and progressively over time; (2) all modules in our model can be trained jointly in an end-to-end manner to be compatible with each other.

Suppose the output appearance feature of last fully-connected layer of convolutional appearance module is $\mathbf{p}_t$ at frame $t$,  then the hidden representation $\mathbf{z}_t$ is calculated by:
\begin{equation}
\mathbf{z}_t = g(\mathbf{W} \mathbf{p}_t + \mathbf{V} \mathbf{z}_{t-1} + b)
\end{equation}
Herein $\mathbf{W}$ and $\mathbf{V}$ are the transformation matrices for appearance feature in current frame and the hidden representation in the previous frame. We use a ReLU function as the activation function $g$ since it eliminates potential vanishing-gradient problems. In practice, we employ two-layer recurrent networks in our recurrent dynamic module, which is expected to potentially learn more latent temporal dynamics than single-layer recurrent networks.

\subsection{Temporal Attention Module}
The attention scheme can be leveraged not only for the selection of the salient facial regions in the spatial domain, but also for the selection of the salient sequential segments (frames) in the temporal domain. Hence we propose to use a temporal attention module on top of the recurrent dynamic module to capture the temporal salience information. The temporal attention module produces an attention weight as the salience score for each frame, thereby filtering the output information flow from the recurrent dynamic module. 

Formally, suppose the output hidden-unit representation of the recurrent dynamic module is $\mathbf{z}_t$ at the frame $t$, then the temporal attention score $e_t$ is modeled by a two-layer perceptron:
\begin{equation}
e_t = \sigma (\mathbf{v}^\top \tanh (\mathbf{M} \mathbf{z}_t + \mathbf{b}) + c)
\label{eqn:temp_att_normalize}
\end{equation}
Here $\mathbf{M}\in \mathbb{R}^{n' \times n}$ is the weight matrix and $\mathbf{b}$ is the bias term for the first perceptron layer, $\mathbf{v} \in \mathbb{R}^{n'}$ is the fusion vector of the second layer. Here $n'$ is a hyper-parameter that is the dimension of transformed mid-representation. Again, we use a sigmoid function to constrain the values between $0$ and $1$. We employ this perceptron to measure the relevance of each frame to the objective task, i.e., age estimation. Next, the attention score is normalized over the whole video to get the final temporal attention weight $o_t$:
\begin{equation}
o_t = \frac{e_t}{\sum_{t'=1}^T e_{t'}}
\end{equation}
The obtained temporal attention weights are used to control how much information for each frame is taken into account to perform the age estimation. Concretely, we calculate the weighted sum of the hidden-unit representation for all frames of the recurrent dynamic module to be the information summary $\mathbf{s}$ for the whole video:
\begin{equation}
\mathbf{s} = \sum_{t=1}^T o_t \mathbf{z}_t
\end{equation}
Ultimately, the predicted age of the corresponding subject involved in the video is estimated by a linear regressor:
\begin{equation}
\tilde{y} = \mathbf{k} \cdot \mathbf{s} + b
\end{equation}
where $\mathbf{k}$ contains the regression weights.

\subsection{End-to-end Parameter Learning}
Given a training dataset $\mathcal{D} = \{ \mathbf{x}_{1,\ldots, T_{(n)}}^{(n)}, y^{(n)} \}_{n = 1, \ldots, N}$ containing $N$ pairs of facial videos and their associated subject's age, we learn the involved parameters of all four modules (convolutional appearance module, spatial attention module, recurrent dynamic module, and temporal attention module) and the final linear regressor jointly by minimizing the mean absolute error loss of the training data:
\begin{equation}
\mathcal{L} = \frac{1}{N} \sum_{n=1}^N |\tilde{y}^{(n)} - y^{(n)}|
\label{eqn:loss}
\end{equation}
Since all modules and the above loss function are analytically differentiable, our whole model can be readily trained in an end-to-end manner. The loss is back-propagated through four modules successively using back-propagation
through time algorithm~\cite{backprop} in the recurrent dynamic module and normal back-propagation way in other parts.

\section{Experimental Setup}

\subsection{Datasets}
\subsubsection{UvA-NEMO Smile Database}
\begin{figure*}[!htb]
 \begin{center}
 \begin{tabular}{cc}
 \includegraphics[width=0.48\linewidth]{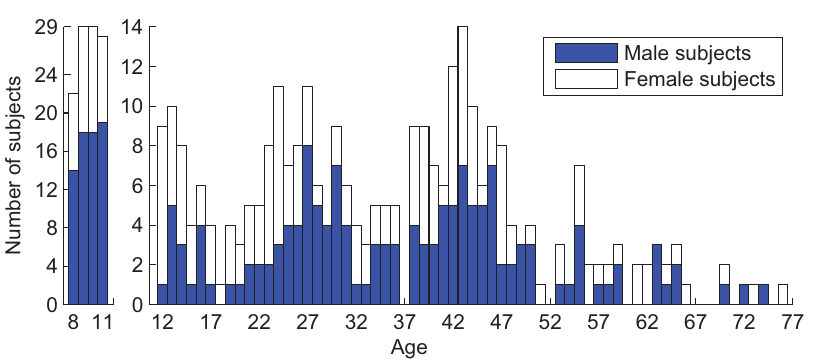} &
  \includegraphics[width=0.48\linewidth]{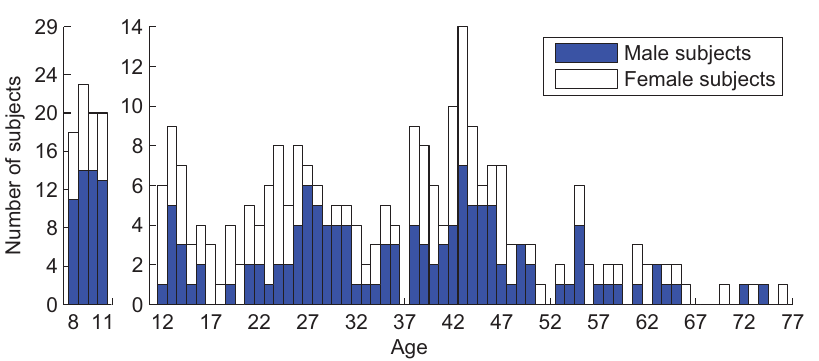}\\
  (a)&(b)\\
\end{tabular}  
 \end{center}
 \vspace{-.1in}
 \caption{Age and gender distributions of the subjects in (a) the UvA-NEMO Smile and \highlight{(b) the UvA-NEMO Disgust databases.}}
 \label{fig:age_dist}
\end{figure*}

We evaluate the performance of our proposed age estimation architecture on the UvA-NEMO Smile Database, which was collected to analyze the temporal dynamics of spontaneous/pose smiles for different ages~\cite{DibekliogluECCV2012}. The database is composed of 1240 smile videos (597 spontaneous and 643 posed) recorded from 400 subjects (185 female and 215 male). The involved subjects span an age interval ranging from 8 to 76 years. Figure~\ref{fig:age_dist}(a) presents the age and gender distribution. 

To collect posed smiles, each subject was asked to pose a smile as realistically as possible.  Spontaneous smile was elicited by short and funny video segments. For each subject, approximately five minutes of recordings were made and the genuine smiles were then segmented. A balanced number of spontaneous and posed smiles are selected and annotated by the consensus of two trained annotators for each subject. Each segment of video starts/ends with neutral or near-neutral expressions.

\subsubsection{\highlight{UvA-NEMO Disgust Database}}
\highlight{To evaluate the proposed approach for other facial expressions, we use the UvA-NEMO Disgust Database~\cite{Hamdi2015} that has been recorded during the acquisition of the UvA-NEMO Smile Database using the same
recording/illumination setup. The database consists of the (posed) disgust expressions of 324 subjects (152 female, 172 male), where 313 of these subjects are also included in the 400 subjects of the UvA-NEMO Smile Database. For each subject, one or two posed disgust expressions were selected and annotated by seeking consensus of two trained annotators. Each of the segmented disgust expressions starts and ends with neutral or near-neutral expressions. The resulting database has 518 posed disgust videos. Subjects' ages vary from 8 to 76 years as shown in Figure~\ref{fig:age_dist}(b).}

\subsection{Tracking and Alignment of Faces}
To normalize face images in terms of rotation and scale, 68 landmarks on facial boundary (17 points), eyes \& eyebrows (22 points), nose (9 points), and mouth (20 points) regions are tracked using a state-of-the-art tracker~\cite{baltruvsaitis2016openface}. The tracker employs an extended version of Constrained Local Neural Fields (CLNF)~\cite{baltrusaitis2013constrained}, where individual point distribution and patch expert models are learned for eyes, lips and eyebrows. Detected points by individual models are then fit to a joint point distribution model. To handle pose variations, CLNF employs a 3D latent representation of facial landmarks. 

The movement of the tracked landmarks is smoothed by the 4253H-twice method~\cite{velleman1980definition} to reduce the tracking noise. Then, each face image (in videos) is warped onto a frontal average face shape using a piecewise linear warping. Notice that the landmark points are in the same location for each of the warped/normalized faces. Such a shape normalization is applied to obtain (pixel-to-pixel) comparable face images regardless of expression or identity variations. The obtained images are cropped around the facial boundary and eyebrows, and scaled so as to have a resolution of $114 \times 114$ pixels. Images are then converted to gray scale.

\subsection{Settings}
Following the experimental setup of Dibeklio\u{g}lu et al.~\cite{Hamdi2015}, we apply a 10-fold cross-validation testing scheme with the same data split to conduct experiments. There is no subject overlap between folds. Each time one fold is used as test data and the other 9 folds are used to train and validate the model. The parameters are optimized independently of test data. For the recurrent dynamic module, the number of hidden units is tuned by selecting the best configuration from the  set $\{128, 256, 512\}$ using a validation set. To prevent  over-fitting, we adopt Dropout~\cite{Dropout} in both the convolutional networks and the recurrent networks and we augment the loss function with L2-regularization terms. Two dropout values, one for the recurrent dynamic module and one for the convolutional appearance module, are validated from the option set $\{0, 0.1, 0.2, 0.4\}$. The L2-regularization parameter $\lambda$ is validated from the option set $\{0, 1e^{-4}, 3e^{-4}, 5e^{-4}, 1e^{-3}, 3e^{-3}, 5e^{-3}\}$. We perform gradient descent optimization using RMSprop~\cite{RMSprop}. The gradients are clipped between $-5$ and $5$~\cite{gradientclip} to avoid potential gradient explosion.

\section{Experiments}
We first investigate the different mechanisms to implement spatial attention and validate the advantages of our proposed spatially-indexed mechanism over other options. Then we present the qualitative and quantitative evaluation on our model respectively, especially to validate the functionality of each module. Subsequently, we compare our model with state-of-the-art methods. Specifically, we make a statistical analysis on predicted error distributions to investigate the difference between the method based on the handcrafted features and our method with automatically learned features. Finally, we evaluate our model on disgust expression to test the generalization of our model to other facial expressions. 

\subsection{Investigation of Spatial Attention Modules}
\label{sec:experiment_spatial_att}
We first conduct experiments to investigate the effectiveness of the proposed spatially-indexed attention mechanism compared to other options for the spatial attentions module. Then we illustrate the effect of position where the spatial attention module is inserted in the convolutional appearance module.
\begin{figure*}[!tb]
	\begin{minipage}[c]{0.5\linewidth} 
   \centering
   \includegraphics[width=0.92\linewidth]{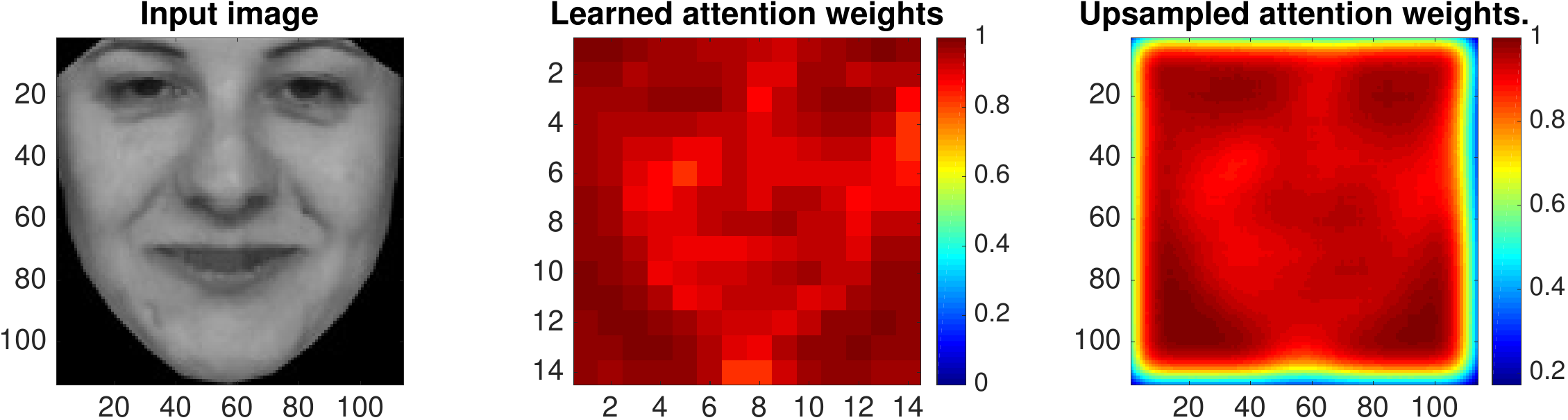} \\
   \minorhighlight{(a)} 
   \end{minipage}
    \begin{minipage}[c]{0.5\linewidth} 
	\centering
   \includegraphics[width=0.92\linewidth]{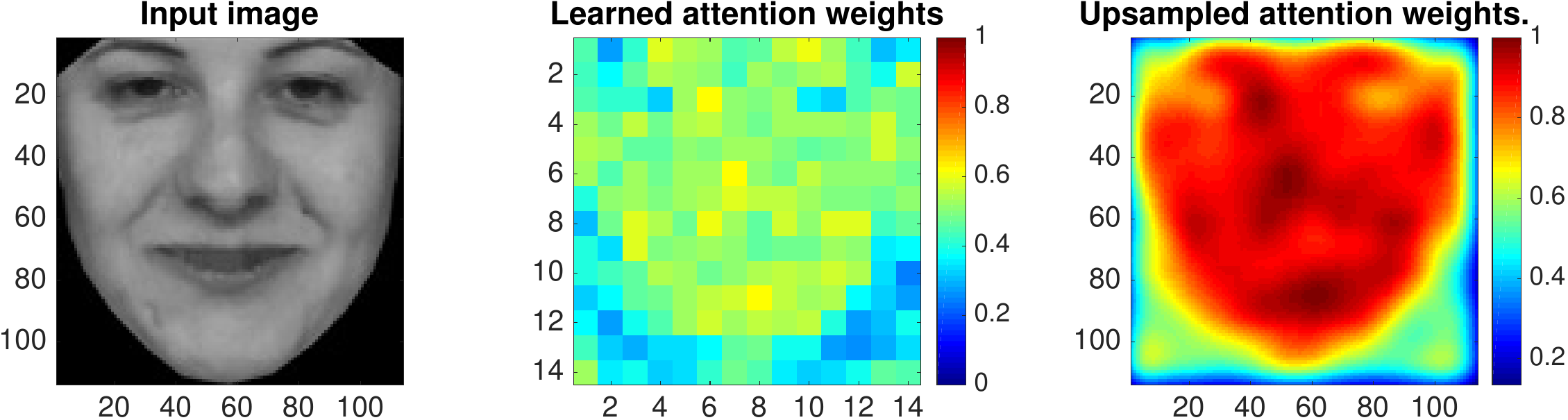} \\
   \minorhighlight{(b)}
   \end{minipage}
   \begin{minipage}[c]{0.5\linewidth} 
   \centering
   \vspace{3mm}
	\includegraphics[width=0.92\linewidth]{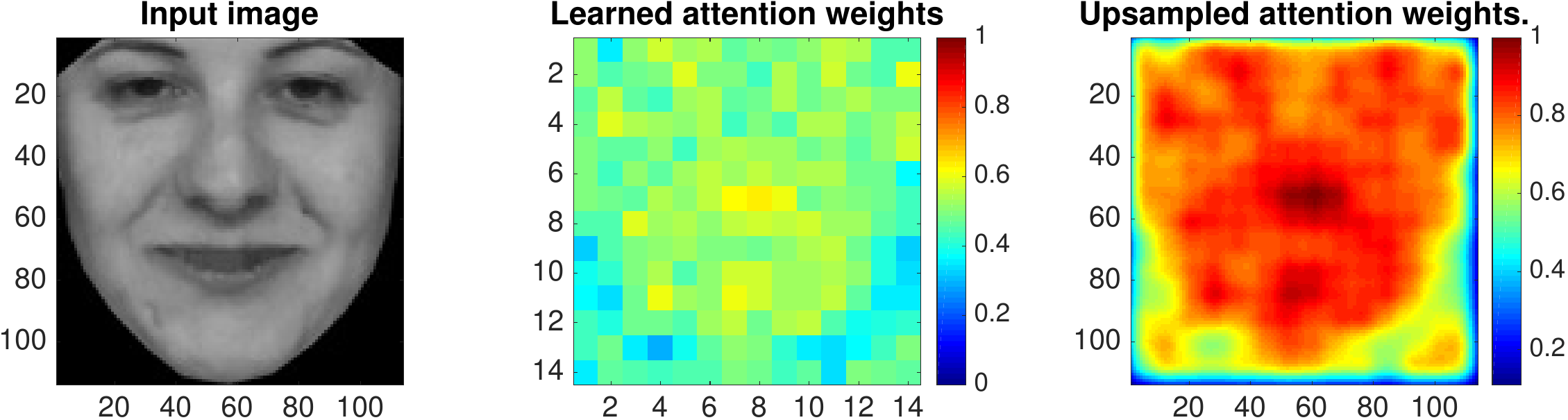} \\
    \minorhighlight{(c)}
	\end{minipage}
   \begin{minipage}[c]{0.5\linewidth} 
   \centering
   \vspace{3mm}
   \includegraphics[width=0.92\linewidth]{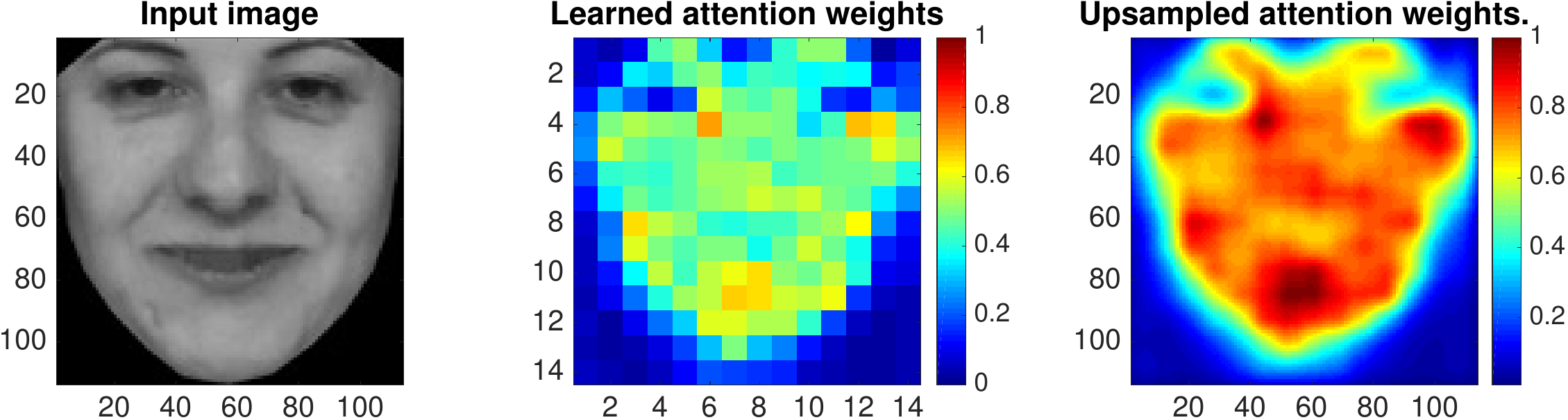} \\
   \minorhighlight{(d)}
   \end{minipage}
	
   \caption{The visualization of the learned attention weights on four different spatial attention mechanisms: \minorhighlight{(a) Spatially-agnostic mechanism, (b) fully spatially-indexed mechanism, (c)  mediate spatially-indexed mechanism, and (d) spatially-indexed mechanism.} For each group of plots corresponding to a mechanism, we first present the original input image, subsequently we visualize the learned weights from the spatial attention module, and finally we up-sample the attention distribution back to the size of input image by a Gaussian filter. Note that the spatial attention module is inserted after the 2nd convolutional layer in this set of experiments.}   
\label{fig:spatial_att_mechanism}
\end{figure*}

\subsubsection{Comparison of different spatial attention mechanisms.}
\label{sec:spatial_attention_mechanism}
We propose a spatially-indexed attention mechanism indicated in Equation~\eqref{eqn:spatial_att1} to model the spatial attention weight $\mathbf{A}$. 
In order to validate the design motivation behind it, we investigate the difference of four different mechanisms:
\begin{itemize}
\item \textbf{Spatially-agnostic} mechanism: both $\mathbf{W}$ in the first layer and $\mathbf{u}$ in the second layer are shared across all entries of the feature map, which is the typical attention model~\cite{Kelvin2015}.
\item \textbf{Fully spatially-indexed} mechanism: both the transformation weights $\mathbf{W}$ and $\mathbf{u}$ are individually designed for each entry of the feature map.
\item \textbf{Mediate spatially-indexed} mechanism: the first layer shares the transformation weights $\mathbf{W}$ while  the second layer model each entry by individual weight $\mathbf{u}$.
\item \textbf{Spatially-indexed} mechanism (adopted by our model): the weight $\mathbf{W}$ of the first layer is individually designed and the weight $\mathbf{u}$ in the second layer is shared. 
\end{itemize}
Figure~\ref{fig:spatial_att_mechanism} presents the qualitative comparison between these four mechanisms. For each group of images, we first visualize the learned spatial attention weights for each option directly (the middle plot of each group), then we up-sample it back to the initial size of the input image by a Gaussian filter (the last plot of each group). This allow us to visualize the receptive field of attention.

It shows that the attention distribution of the spatially-agnostic mechanism appears blurred and less contrasting than other mechanisms. It is only able to capture the salient regions around mouth and near eyebrow. It even gives high scores to the background. It is not surprising since the receptive fields of each entry overlap, hence it is hard for the shared weights to capture the fine-grained differences. The fully spatially-indexed mechanism can roughly capture the contour of the facial regions in the images but with no highlights inside the facial area. This is because individual weights can hardly model the spatial continuity in the face. In contrast, the spatially-indexed mechanism achieves the best result among all options. Furthermore, adopting shared weights in the first layer and individual weights in the second layer (mediate spatially-indexed mechanism) is much worse than the other order. It is probably because the individual weights can hardly take effect after the smoothing by the shared weights. 
Therefore, our model employs the spatially-indexed mechanism, which can not only clearly distinguish the face from the background, but also capture the salient regions like the area under the eye, area around mouth and two nasolabial folds in cheeks. More examples are presented in Section~\ref{sec:spatial_att_example}. \highlight{Quantitative comparison is also provided in Table~\ref{table:attention_mechanims}, which demonstrates that the spatially-indexed mechanism outperforms other spatial attention mechanisms.}

\begin{table}[!tb]
\caption{\highlight{Mean absolute error (years) for four different spatial attention mechanisms on the UvA-NEMO Smile Database.}}
\vspace{-.1in}
\centering
\vspace{2mm}
\renewcommand\arraystretch{1.4}
\resizebox{0.75\linewidth}{!}{
\begin{tabular}{l|c}
\Xhline{1.0pt}
 \highlight{\textbf{Attention Mechanism}} & \highlight{\textbf{MAE (years)}}\\
 \hline
 $\highlight{\text{Spatially-agnostic}}$ & $\highlight{4.95}$ $\highlight{(\pm 5.77)}$\\
 $\highlight{\text{Fully spatially-indexed}}$& $\highlight{4.94}$ $\highlight{(\pm 5.65)}$\\
 \highlight{\text{Mediate spatially-indexed}}& $\highlight{5.01}$ $\highlight{(\pm 5.45)}$\\
 \highlight{\text{Spatially-indexed}}& $\highlight{4.74}$ $\highlight{(\pm 5.70)}$\\
\Xhline{1pt}
\end{tabular}
}
\label{table:attention_mechanims}
\end{table}

\subsubsection{The effect of the position of the spatial attention module} 
\label{sec:effect_att_position}
\begin{figure}[!tb]
 \begin{center}
\includegraphics[width=0.95\linewidth]{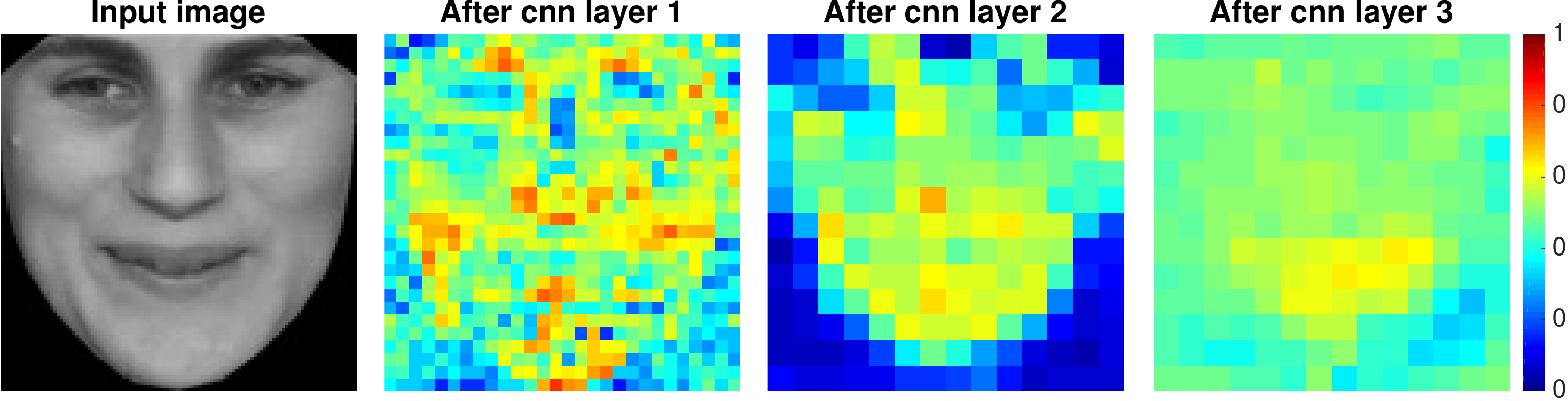}
 \end{center}
  \vspace{-.1in}
 \caption{The visualization of the learned spatial attention weights when placing the spatial attention module after different convolutional layers. Note that we adopt the spatially-indexed mechanism described in Section~\ref{sec:spatial_attention_mechanism}.}
 \label{fig:spatial_position}
\end{figure}

Theoretically, the spatial attention module can be placed after any convolutional layer in the appearance module. However, the latter convolutional layers output feature maps with larger receptive fields for each entry than the previous layers, which leads to more overlapping receptive fields of adjacent entries. As a result, each spatial attention weight also corresponds to a larger receptive field in the input image. Figure~\ref{fig:spatial_position} shows the learned spatial attention weights after inserting the spatial attention module after different convolutional layers. In the case that the spatial attention is placed after the first layer, the distribution of learned attention weights is very noisy. This is because the small receptive fields of each attention weight results in excessively fine-grained modeling, which causes over-fitting. In contrast, placing the spatial attention module after the third convolutional layer generates more coarse and less contrasting attention weight distributions, which weakens the effect of the spatial attention module. We achieve a good balance by inserting the spatial attention module after the second convolutional layer, as shown in Figure~\ref{fig:spatial_position}.

\subsubsection{\highlight{Investigation of internal scheme of spatial attention module}}
\highlight{To investigate whether the first layer of the spatially-indexed attention module is able to capture the local detailed variation as we expect, we attempt to make a visualization on the learned weights of the first attention layer. Since the output of the first attention layer in Equation~\ref{eqn:spatial_att1} is a 3-D tensor with dimensions $M \times N \times d $, i.e., each pixel in the feature map of size $M \times N$ is a $d$-dim vector, it is hard to visualize the learned weights directly. Instead, we first perform Principal Component Analysis (PCA) on the learned attention weights, then we visualize the first 3 principle components which accounts for $55 \%$ the variance.}  

\begin{figure}[!tb]
 \begin{center}
 \includegraphics[width=0.95\linewidth]{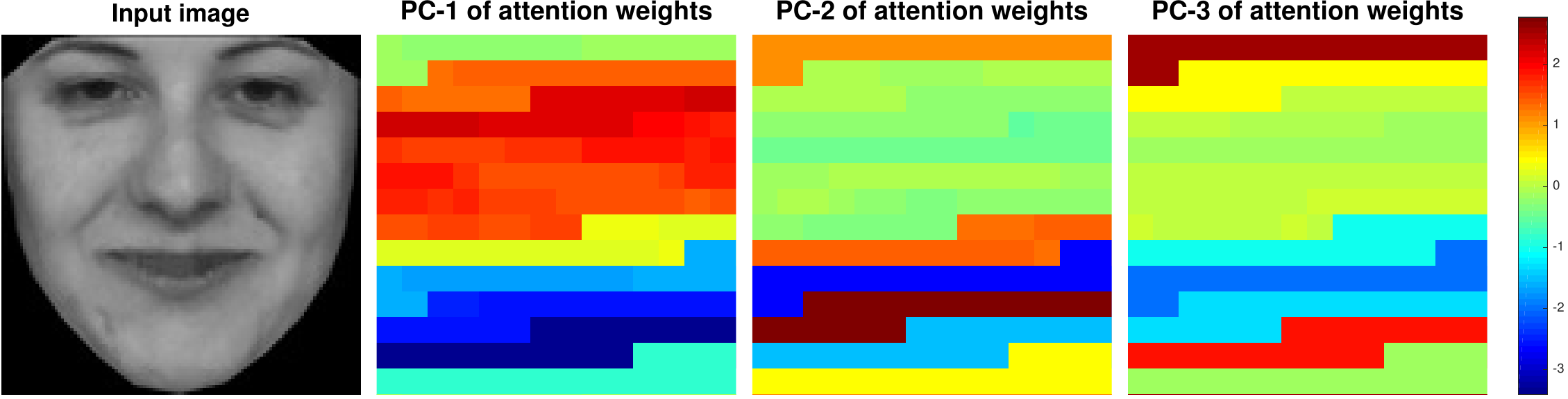}
 \end{center}
  \vspace{-.1in}
 \caption{\highlight{The visualization of the learned attention weights of the first spatially-indexed attention layer. The Principal Component Analysis (PCA) is performed on the 256-dimension attention weights for each pixel of the attention map. The first 3 principle components (PC) which accounts for 55\% variance are visualized.}}
 \label{fig:spatial_pos_vis}
\end{figure}

\highlight{As shown in Figure~\ref{fig:spatial_pos_vis}, the first layer of the spatially-indexed attention module tends to learn the sharp variance vertically in the feature map. The whole face is roughly divided into several vertical blocks; for instance, the PC-1 of attention weights focuses more on the area around eyes and eyebrows. Resulting vertical clusters of attention scores may be explained by the horizontally-symmetric appearance of faces.}

\subsection{Quantitative Evaluation of Functionality of each Module}
\begin{figure}[!t]
 \begin{center}
 \includegraphics[width=0.75\linewidth]{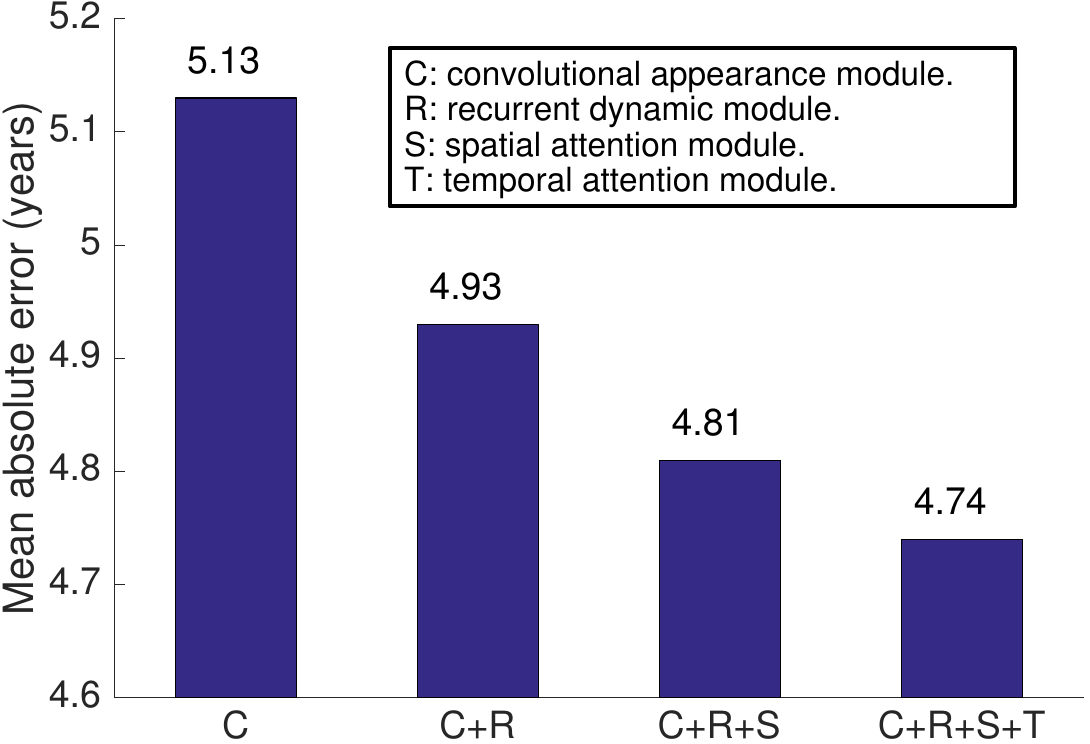} 
 \end{center}
  \vspace{-.1in}
 \caption{Mean absolute error (years) for different functional modules on the UvA-NEMO smile database.}
 \label{fig:module_function}
\end{figure}

Next we perform the quantitative evaluation on the four functional modules of our module, by evaluating the functionality of each module and the contribution it makes to performance of the whole architecture. To this end, we conduct ablation experiments which begin with the single convolutional appearance module in the system and then incrementally augments the system by one module at a time. When we only employ the convolutional appearance module (without modeling dynamics) in our system, we perform age estimation on each single image in the video and then average the predicted results as the final age estimation. Figure~\ref{fig:module_function} presents the performance of all ablation experiments. 

The individual convolutional appearance module in the system achieves an age estimation performance of $5.13$ years' mean absolute error (MAE), which is an encouraging result considering the fact that it only performs appearance learning (without any dynamics information involved for age estimation). More detailed comparisons to handcrafted features are made in Table~\ref{table:inter_compare2} presented in subsequent Section~\ref{sec:inter_comparision}. Equipping the system with the recurrent dynamic module results in $4.93$ years' MAE, which indicates that the dynamics learning by recurrent dynamic module makes a substantial improvement. Subsequently, the spatial attention module is added into the system to capture the spatial salience in each facial image, and the MAE is decreased to $4.81$ years. 
We will present a qualitative visualization of learned spatial attention salience in Section~\ref{sec:spatial_att_example} and Figure~\ref{fig:spatial_att_vis}. 
Finally, including the temporal attention module, leading to our full end-to-end system, results in the best performance with 4.74 years' MAE.

It should be mentioned that the power of the temporal attention module is actually not fully exploited, since this data has been segmented to retain the smile phase mostly. Most of irrelevant segments before and after the smile have been removed. The temporal attention module will be shown qualitatively to be capable of capturing the key smile segment precisely in Section~\ref{sec:temporal_att} and Figure~\ref{fig:temporal_att}. Hence more improvement by the temporal attention module is expectable given temporally noisier data. 

\subsection{\highlight{Effect of Temporal Phases}}
\highlight{A facial expression consists of three non-overlapping temporal phases, namely: 1) the onset (neutral to expressive), 2) apex, and 3) offset (expressive to neutral), respectively. 
To understand the informativeness of different temporal phases in age estimation, we evaluate the accuracy on each temporal phase using our model pre-trained on the whole expression (from the beginning of onset to the end of offset).}

\highlight{As shown in Table~\ref{table:temporal_phase}, the apex phase contributes most to the task of age estimation. It is reasonable since smiling faces tend to contain more discriminative features like wrinkles than other phases. It is also consistent with the distribution of temporal attention weights showed in Section~\ref{sec:temporal_att} and Figure~\ref{fig:temporal_att}. As expected, the combined use of consecutive phases leads to a better performance since these phases have different temporal dynamics and appearance patterns that can provide additional information. Therefore, we may claim that our approach can still be used (with a relative decrease in accuracy) even if the whole duration of an expression is not captured.}

\begin{table}[!tb]
\caption{\highlight{Mean absolute error (years) for different temporal phases for age estimation on the UvA-NEMO Smile Database.}}
\vspace{-.1in}
\centering
\vspace{2mm}
\renewcommand\arraystretch{1.4}
\resizebox{0.6\linewidth}{!}{
\begin{tabular}{l|c}
\Xhline{1.0pt}
 \highlight{\textbf{Temporal Phase(s)}} & \highlight{\textbf{MAE (years)}}\\
 \hline
\highlight{Onset} 	& $\highlight{8.41}$ $\highlight{(\pm 8.95)}$\\
\highlight{Apex}	& $\highlight{5.91}$ $\highlight{(\pm 7.05)}$\\
\highlight{Offset}	& $\highlight{8.75}$ $\highlight{(\pm 9.68)}$\\
\highlight{Onset~+~Apex} & $\highlight{5.04}$ $\highlight{(\pm 5.80)}$\\
\highlight{Apex~+~Offset}& $\highlight{5.13}$ $\highlight{(\pm 5.83)}$\\
\highlight{Onset~+~Apex~+~Offset}& $\highlight{4.74}$ $\highlight{(\pm 5.70)}$\\
\Xhline{1pt}
\end{tabular}
}
\label{table:temporal_phase}
\end{table}

\subsection{Comparison to Other Methods }
\label{sec:inter_comparision}
	
\begin{table*}[!tb]
\caption{Mean absolute error (years) for different methods on the UvA-NEMO smile database.}
\vspace{-.1in}
\centering
\vspace{2mm}
\small
\renewcommand\arraystretch{1.4}
\resizebox{1.0\linewidth}{!}{
\begin{tabular}{l|l|ccccccccC{0.1pt}c}
\Xhline{1.0pt}
 \multicolumn{2}{c|}{\multirow{2}{*}{\textbf{Method}}} & \multicolumn{10}{c}{\textbf{MAE (years) for Different Age Ranges}}\\
\cline{3-12}
\multicolumn{2}{c|}{} & 0-9 & 10-19 & 20-29 & 30-39 & 40-49 & 50-59 & 60-69 & 70-79 & \multirow{11}{*}{\begin{tikzpicture}
\draw[dotted][line width=0.25mm, black ]
  (0,0) -- (0,7.4);
\end{tikzpicture}} &  All \\
\hline
\multirow{2}{*}{\tabincell{l}{Spatio-temporal} }
& $\text{VLBP}$~\cite{Hadid2011} & 10.69& 12.95& 15.99&18.54 &18.43 &16.58 & 23.80&26.59 & & $15.70$ $(\pm 12.40)$ \\
& $\text{LBP-TOP}$~\cite{LBP-TOP} &9.71 & 11.01& 14.19&15.88 &16.75 &15.29 & 19.70& 23.71&  & $13.83$ $(\pm 10.97)$\\
\hline
\multirow{2}{*}{\tabincell{l}{Dynamics} }
& $\text{Deformation}$~\cite{Hamdi2015} &4.85 &8.72 &12.22 &13.06 & 13.53&11.55 & 14.13& 17.82 & & $10.81$ $(\pm 8.85)$ \\
& $\text{Displacement}$~\cite{DibekliogluICM2012}&5.42 & 9.67& 11.98& 14.53& 12.77& 15.42& 20.57& 20.35&  &$11.54$ $(\pm 11.49)$\\
\hline
\multirow{3}{*}{\tabincell{l}{Appearance} }
& \highlight{$\text{IEF, Neutral}$~\cite{Hamdi2015}}&\highlight{3.27}  & \highlight{3.96}  & \highlight{4.86} &\highlight{6.11} & \highlight{5.53} & \highlight{8.24} & \highlight{12.89} & \highlight{13.35} && \highlight{$5.21$ $(\pm 4.48)$}\\
& $\text{IEF, Fusion}$~\cite{Hamdi2015}& 3.54& 4.38 &5.43 &6.74 &6.01 &8.96 &13.52 &14.05 && $5.71$ $(\pm 4.65)$  \\
& \highlight{$\text{LBP, Neutral}$~\cite{Hamdi2015}}& \highlight{3.86}  & \highlight{4.65}  & \highlight{5.87} & \highlight{6.82} & \highlight{5.58} & \highlight{8.32} & \highlight{9.84} & \highlight{11.9} && \highlight{$5.67$ $(\pm 4.97)$} \\
& $\text{LBP, Fusion}$~\cite{Hamdi2015}& 4.18& 4.99	&6.31 &7.37	&6.19 &8.67	&10.34 &12.93 && $6.12$ $(\pm 5.11)$  \\
& $\text{CNNs}$& 2.30& 3.09& 4.69& 5.26& 5.82& 10.58& 17.20&23.29 & & $5.13$ $(\pm 5.68)$\\
\hline
\multirow{3}{*}{\tabincell{l}{Appearance+Dynamics} }
& $\text{IEF+Dynamics}$~\cite{Hamdi2015}& 3.96& 4.45& 4.50& 5.29& 4.74& 6.85& 12.43& 11.94& & $5.00$ $(\pm 4.25)$\\
& $\text{LBP+Dynamics}$~\cite{Hamdi2015}& 3.49& 4.68& 5.13& 5.85&5.24 & 7.05& 12.17& 12.00& & $5.29$ $(\pm 4.36)$\\
& $\text{SIAM (Our model)}$& 1.79& 2.45& 4.26& 4.97& 5.36& 11.86&16.43 &23.12 & & $\textbf{4.74}$ $(\pm 5.70)$ \\
\Xhline{0.75pt}
 &  Number of Samples & 158 & 333 & 215 & 171 & 250 & 66 & 30 & 17& & 1240\\
 \Xhline{1.0pt}
\end{tabular}
}
\label{table:inter_compare2}
\end{table*}

Next, the model is compared with existing methods for age estimation, that can be applied to sequential images (videos). According to the mechanism of the feature design, these baseline methods can be classified into four categories as listed in Table~\ref{table:inter_compare2}:
\begin{itemize}
\item \textbf{Spatio-temporal}: Hadid et al.~\cite{Hadid2011} propose to employ the spatio-temporal information for classifying age intervals. Particularly, they extract  volume LBP (VLBP) features and feed them to a tree of four SVM classifiers.  Another method using spatio-temporal information is proposed to extract the LBP histograms from Three Orthogonal Planes (LBP-TOP): XY (two spatial dimensions in a single image), XT (X dimension in the image and temporal space T) and YT~\cite{LBP-TOP}. Thus the information in temporal domain is utilized together with information in spatial image domain. These two spatio-temporal methods are implemented for age estimation as baselines by Dibeklio{\u g}lu et al.~\cite{Hamdi2015}, from where we report the results.
\item \textbf{Appearance + Dynamics}: Dibeklio{\u g}lu et al.~\cite{Hamdi2015} is the first study which leverages both the facial dynamics and appearance information for age estimation. They propose several handcrafted dynamics features specifically for facial expressions and combine them with appearance features to perform age estimation through a hierarchical architecture. They combine their dynamics features with four different kinds of appearance descriptors in their system. Among them we select two combinations with the best performance as our baselines: dynamics + IEF (Intensity-based encoded aging features~\cite{IEF}) and dynamics + LBP (local binary patterns)~\cite{Ojala_LBP}.
\item \textbf{Dynamics}: We incorporate the baseline models using sole dynamics information. Following Dibeklio{\u g}lu et al.~\cite{Hamdi2015}, we compare the deformation-based features and the displacement dynamics features~\cite{DibekliogluICM2012}.
\item \textbf{Appearance}: We also compare our method to appearance-based approaches that solely employ IEF and LBP features~\cite{Hamdi2015}, \highlight{where (1) the neutral version estimates the age on the first frame of a smile onset (neutral face); (2) the fusion version averages age estimations from the first and the last frame of a smile onset (a neutral and an expressive face, respectively) for the final prediction.} Furthermore, we evaluate a modified version of our method that uses only convolutional appearance module. 
\end{itemize}

\smallskip\subsubsection{Performance Comparison}
Table~\ref{table:inter_compare2} shows the mean absolute errors (MAE) obtained by four categories of age estimation methods mentioned before. Our model achieves the best performance considering all the age ranges with the minimum MAE of $4.74$ years. While spatio-temporal methods perform worst, the methods utilizing both appearance and dynamics are more accurate than the methods based on sole appearance or dynamics. It illustrates the importance of both appearance and dynamics to the task of age estimation.  

\begin{figure}[!t]
 \begin{center}
 \includegraphics[width=0.72\linewidth]{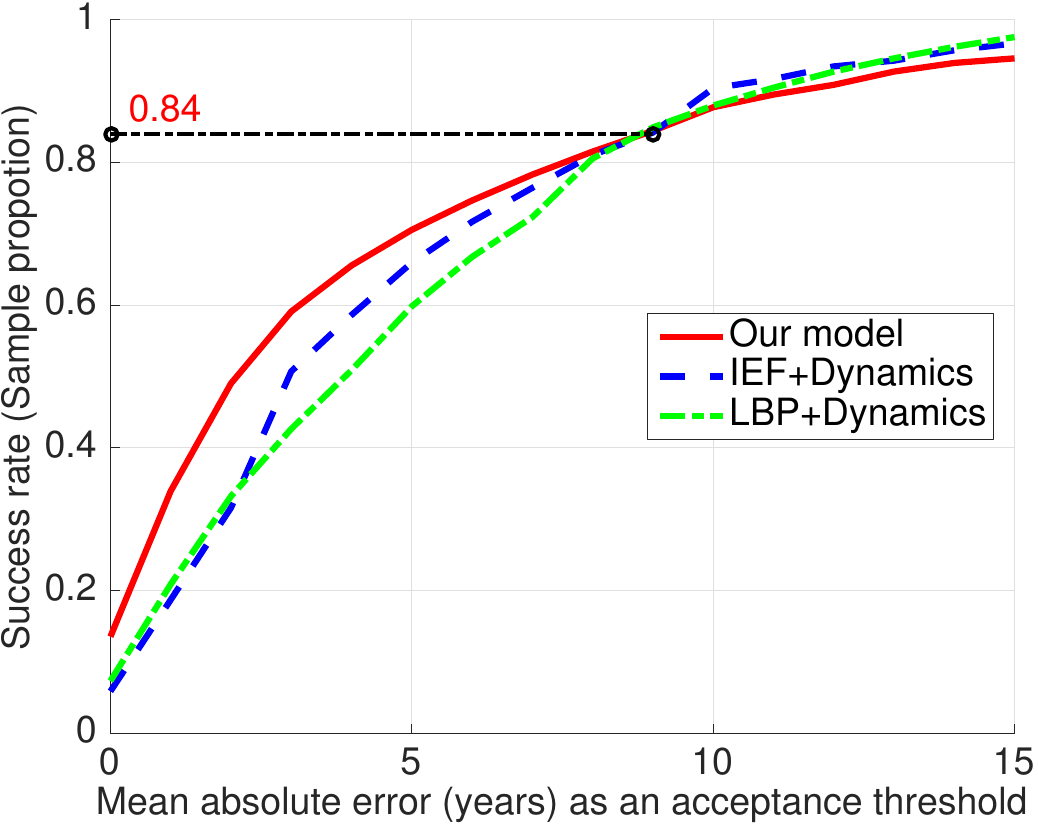} 
 \end{center}
 \vspace{-.1in}
 \caption{Cumulative distribution of the mean absolute error for different models using both appearance and dynamics on the UvA-NEMO Smile Database.}
 \label{fig:cumulative_distribution}
\end{figure}

In particular, the performance of our model is better than the other two methods using both appearance and dynamics: \emph{IEF+Dynamics} and \emph{LBP+Dynamics}~\cite{Hamdi2015}. Figure~\ref{fig:cumulative_distribution} shows the success rate as a function of the MAE for these three methods. 
For age deviations up to nine years, our model outperforms the other two methods. For larger age deviations, the model is slightly worse. Our model suffers from some severe estimation errors on a few samples. These samples appear to be from the high-age groups, where our model is severely hampered by the low sample size in these groups.  
Figure~\ref{fig:age_dist} and Table~\ref{table:inter_compare2} both show that the number of samples in these age ranges is much less than that in the younger age ranges. Compared to handcrafted features used in \emph{IEF+Dynamics} and \emph{LBP+Dynamics}~\cite{Hamdi2015}, the convolutional appearance module and recurrent dynamic module in our model require more training data. 

\begin{figure}[!hbt]
 \begin{center}
 \includegraphics[width=0.75\linewidth]{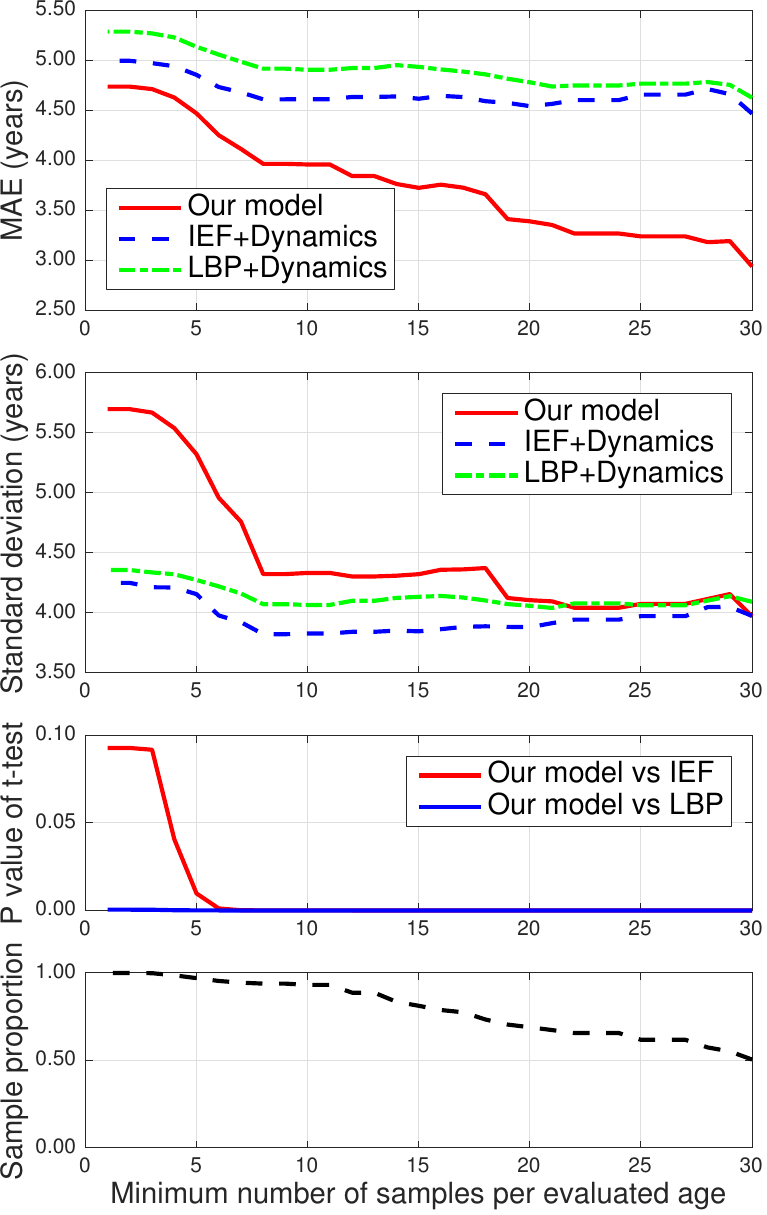} 
 \end{center}
 \vspace{-.1in}
 \caption{Performances of three methods using both the appearance and dynamics on the UvA-NEMO Smile Database as a function of minimum number of samples per evaluated age.}
 \label{fig:MAE_threshold}
\end{figure}

\smallskip\subsubsection{The Effect of Training Size per Age}
\label{sec:train_size}
To investigate the effect of training size per age on the performance of three \emph{Appearance+Dynamics} models, we conduct experiments where the data are reduced by removing ages for which a low number of samples are available.
The results are presented in Figure~\ref{fig:MAE_threshold}. The experiments begin with the case that all the samples involved (threshold $= 1$). As the threshold (minimum number of samples) is increased, 
the performance gap between our model and the other methods becomes larger:
the MAE of our model decreases much faster than the other two methods and the variance also drops deeply at the beginning of the curve. A t-test shows that our model significantly outperforms other methods when the threshold on the number of sample is larger than 5 ($p<0.01$). They are actually quite encouraging results, since these results in turn indicate that larger data tend to explore more potential of our model and make it more promising than the other two methods on the task of age estimation.

\subsection{Qualitative Evaluation of Attention Modules}
\label{sec:qulitative}
In this section, we qualitatively evaluate the spatial attention module and temporal attention module by visualizing the learned attention weights.   
\subsubsection{Spatial Attention Module}
\label{sec:spatial_att_example}
\begin{figure}[!htb]
\begin{center}
\begin{tabular}{cccccccccc}
   \includegraphics[width=0.95\linewidth]{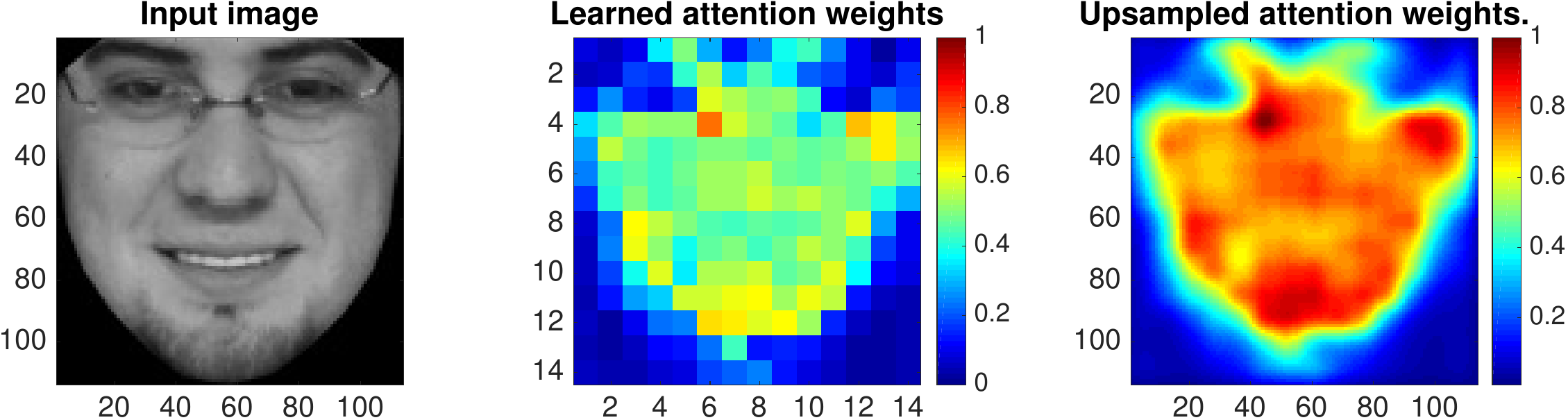} \\
   \includegraphics[width=0.95\linewidth]{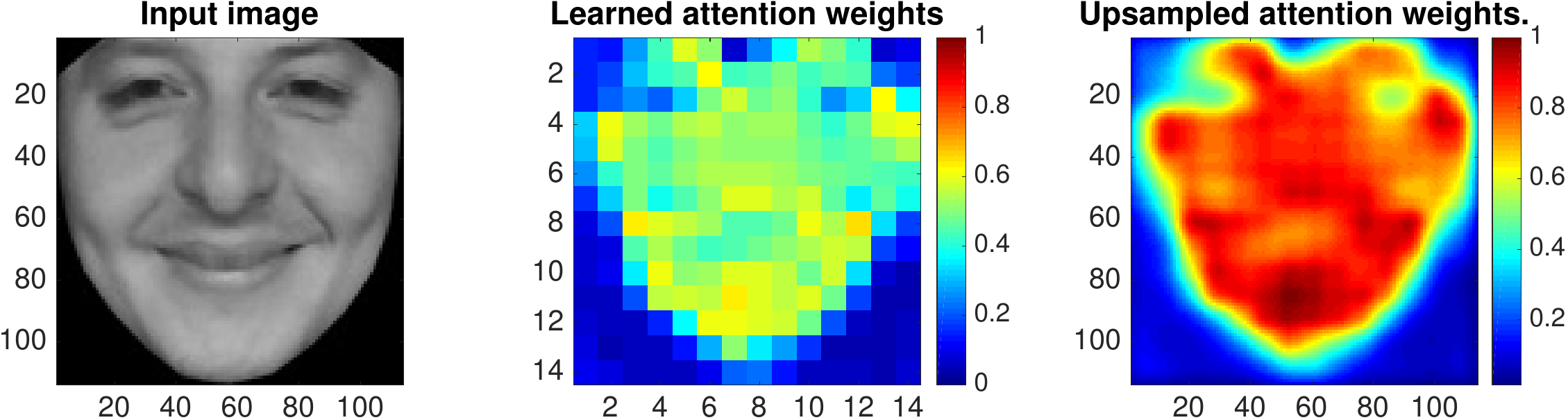} \\
   \includegraphics[width=0.95\linewidth]{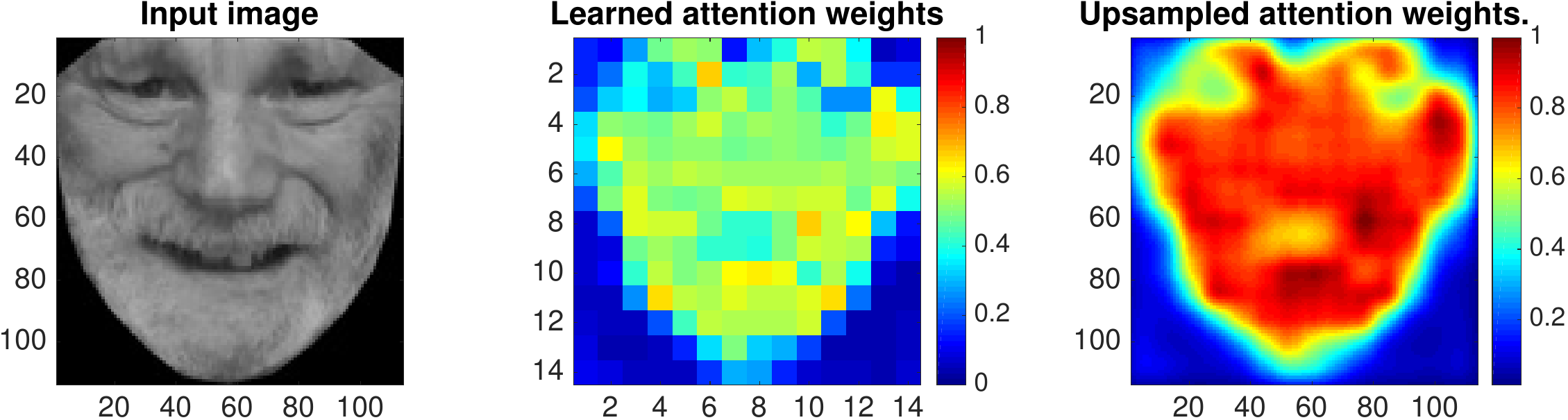} \\
     \vspace{0.2cm}\\
\end{tabular}
\end{center}
\vspace{-.25in}
   \caption{The heat map visualization of the learned attention weights by our spatial attention module. For each subject, the middle plot corresponds to attention weights and the last plot is the up-sampled attention weight distribution back to the size of initial input image by a Gaussian filter.}
\label{fig:spatial_att_vis}
\end{figure}

\begin{figure*}[!tb]
\begin{center}
\resizebox{1.0\linewidth}{!}{
\begin{tabular}{cccccccccc}
   \includegraphics[width=0.1\linewidth]{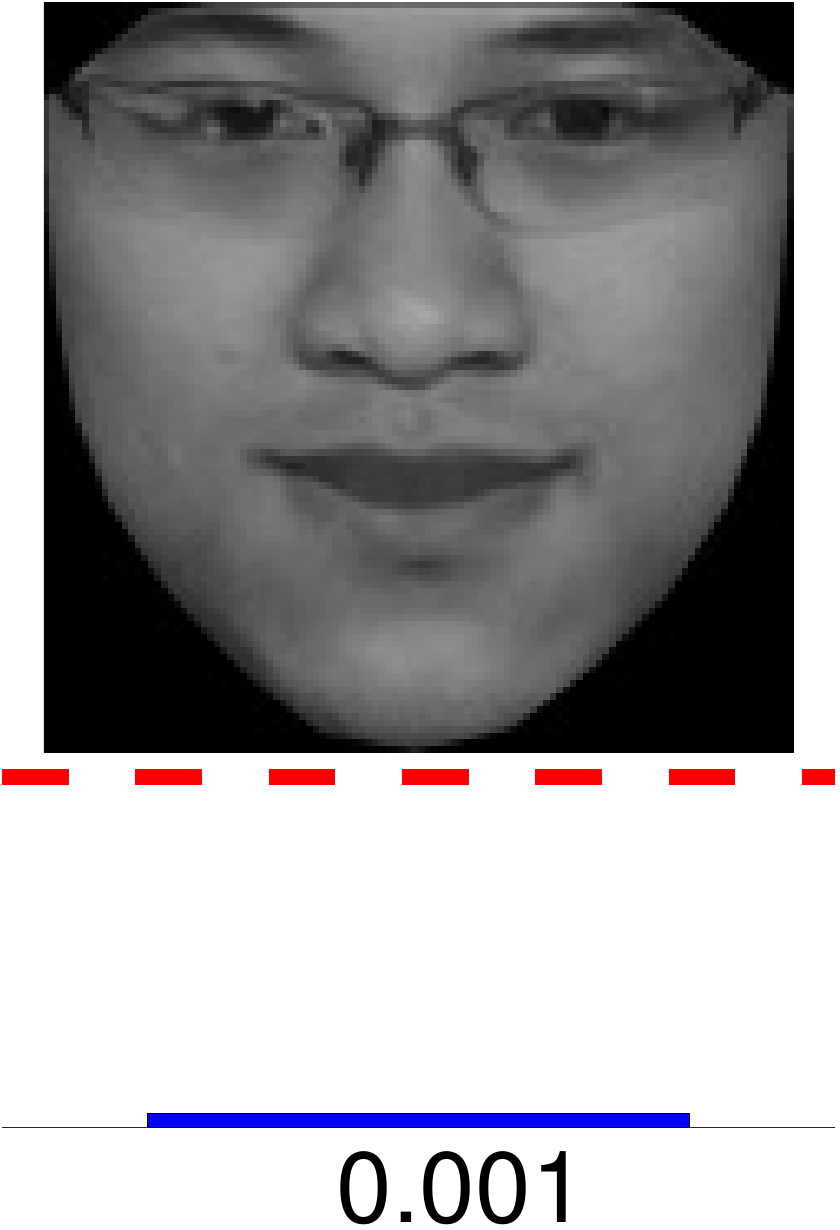} 
   \includegraphics[width=0.1\linewidth]{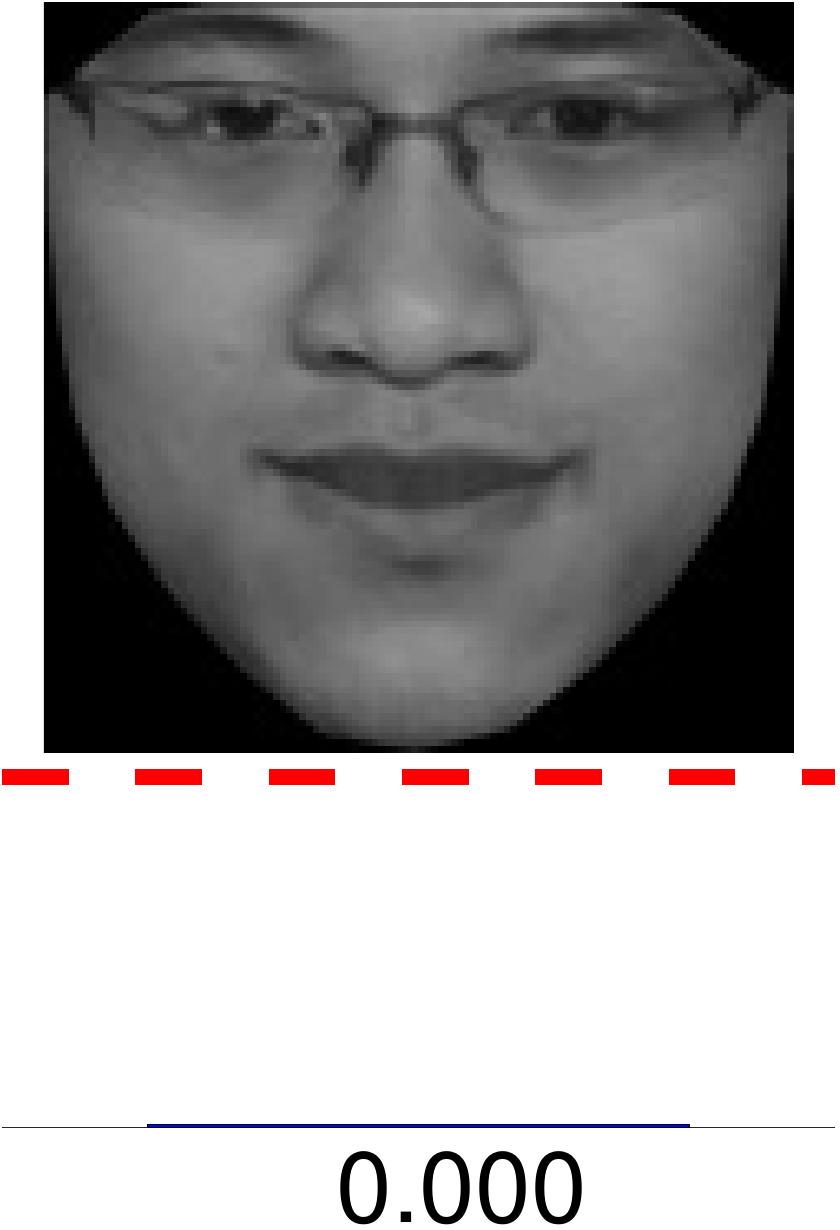}   
   \includegraphics[width=0.1\linewidth]{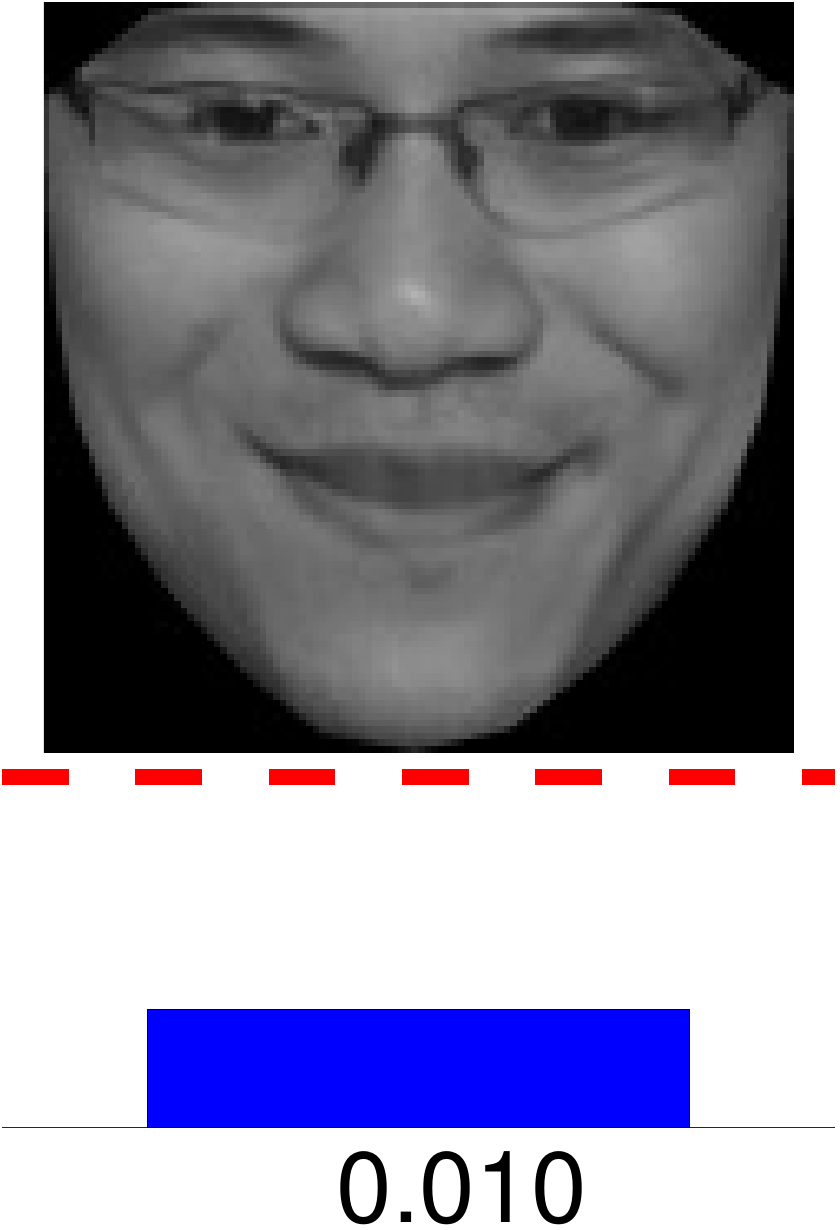}   
   \includegraphics[width=0.1\linewidth]{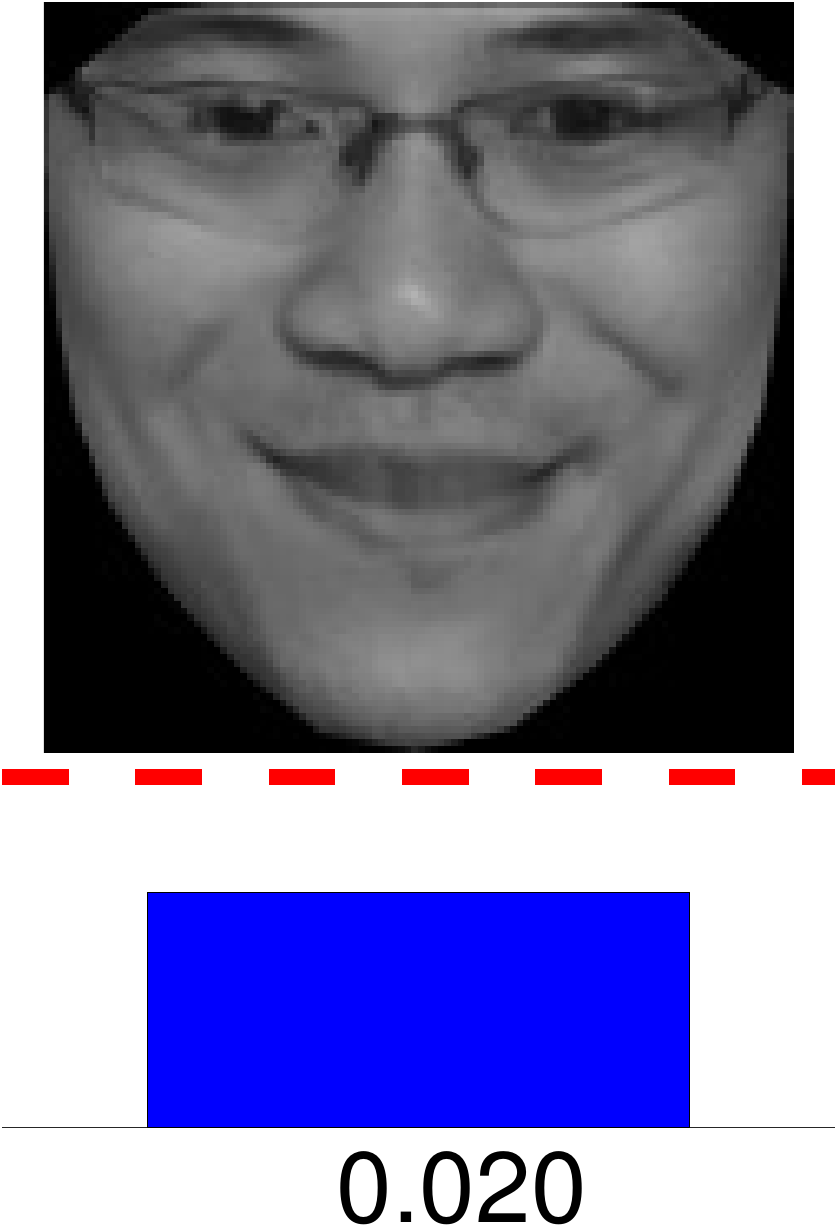}   
   \includegraphics[width=0.1\linewidth]{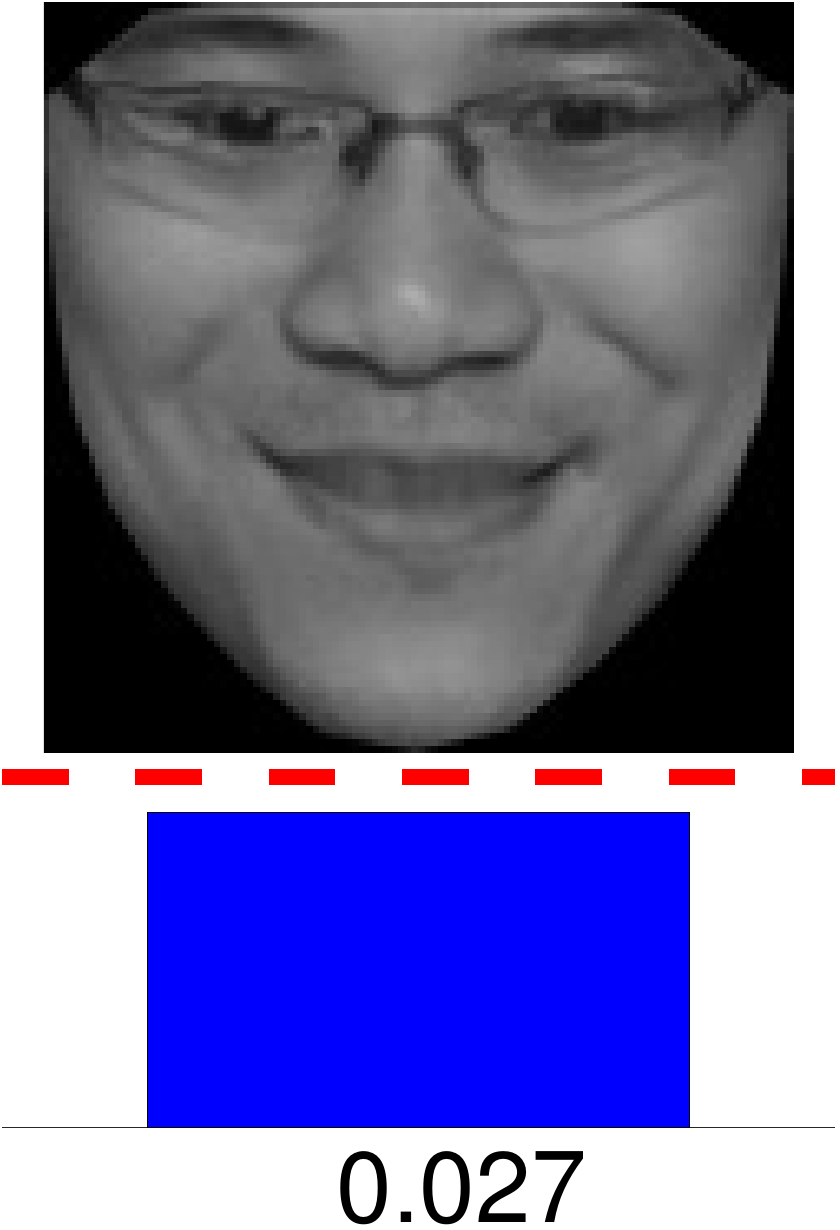}    
   \includegraphics[width=0.1\linewidth]{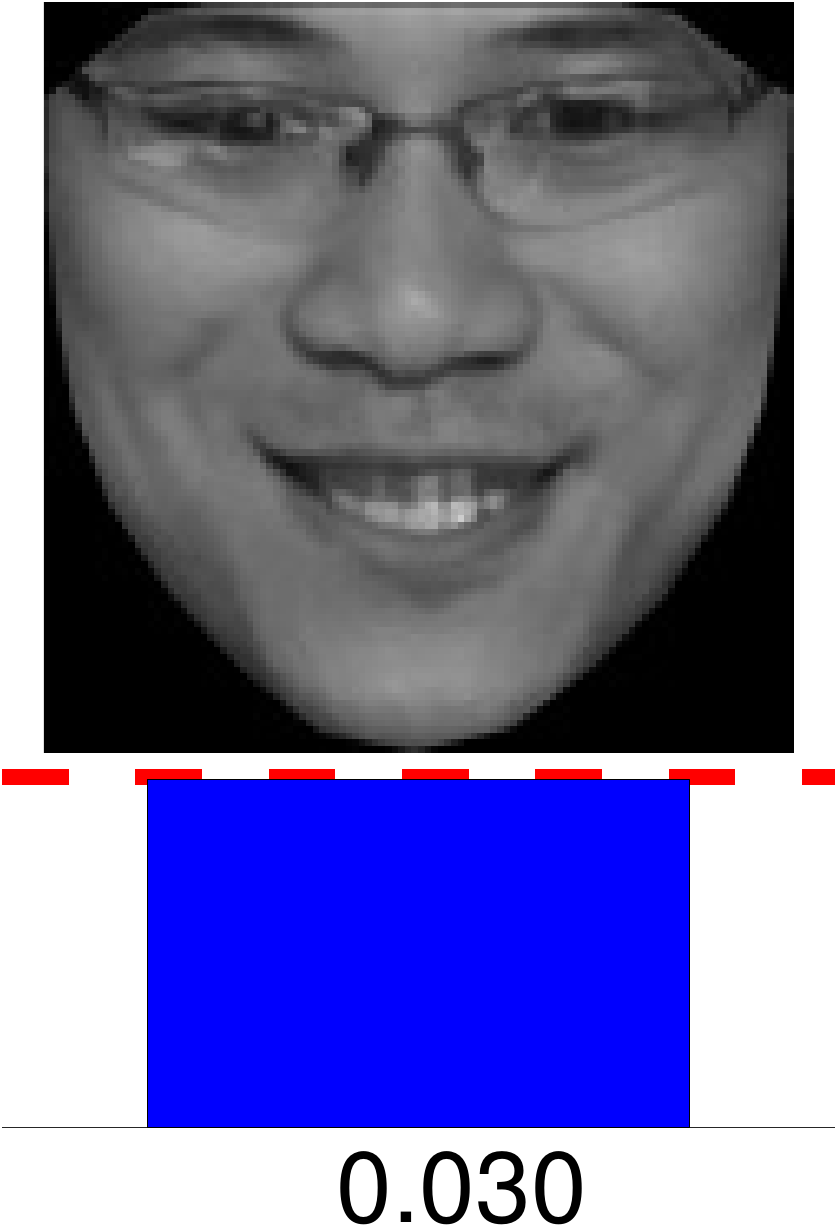}  
   \includegraphics[width=0.1\linewidth]{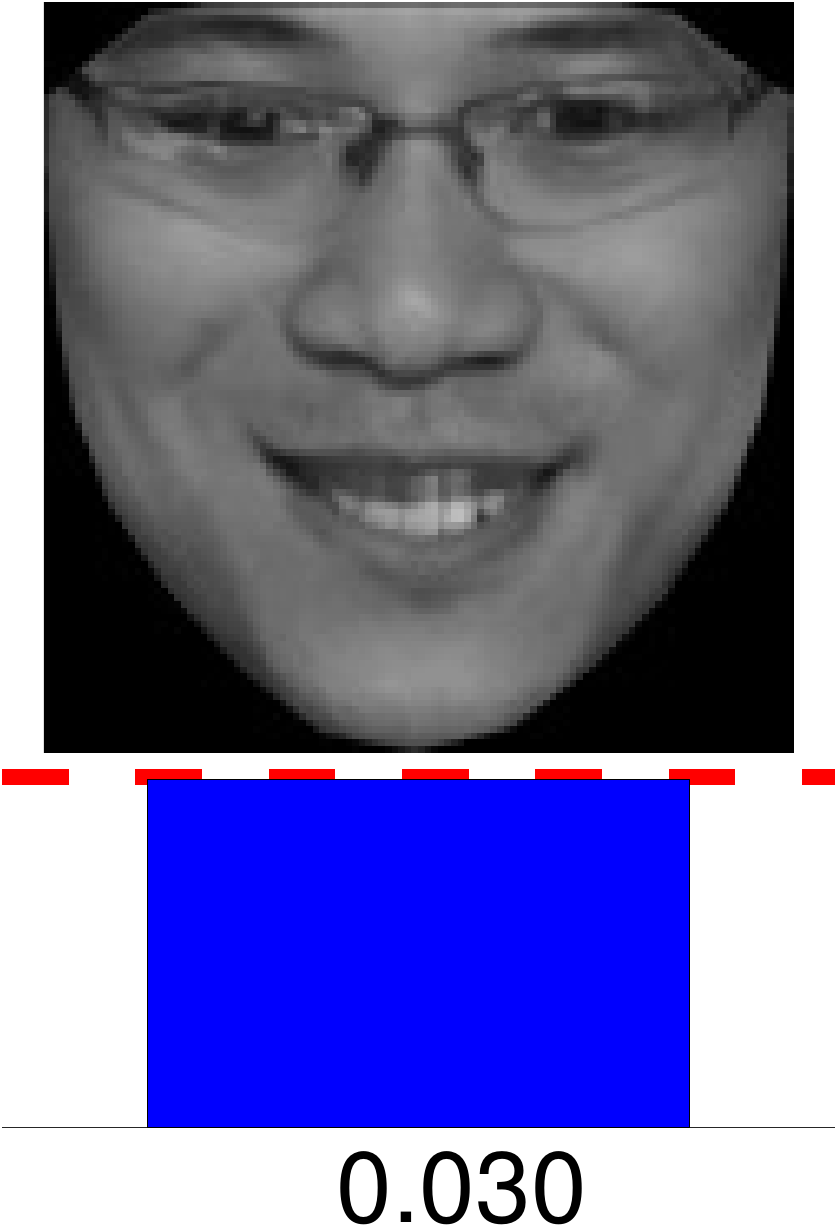}    
   \includegraphics[width=0.1\linewidth]{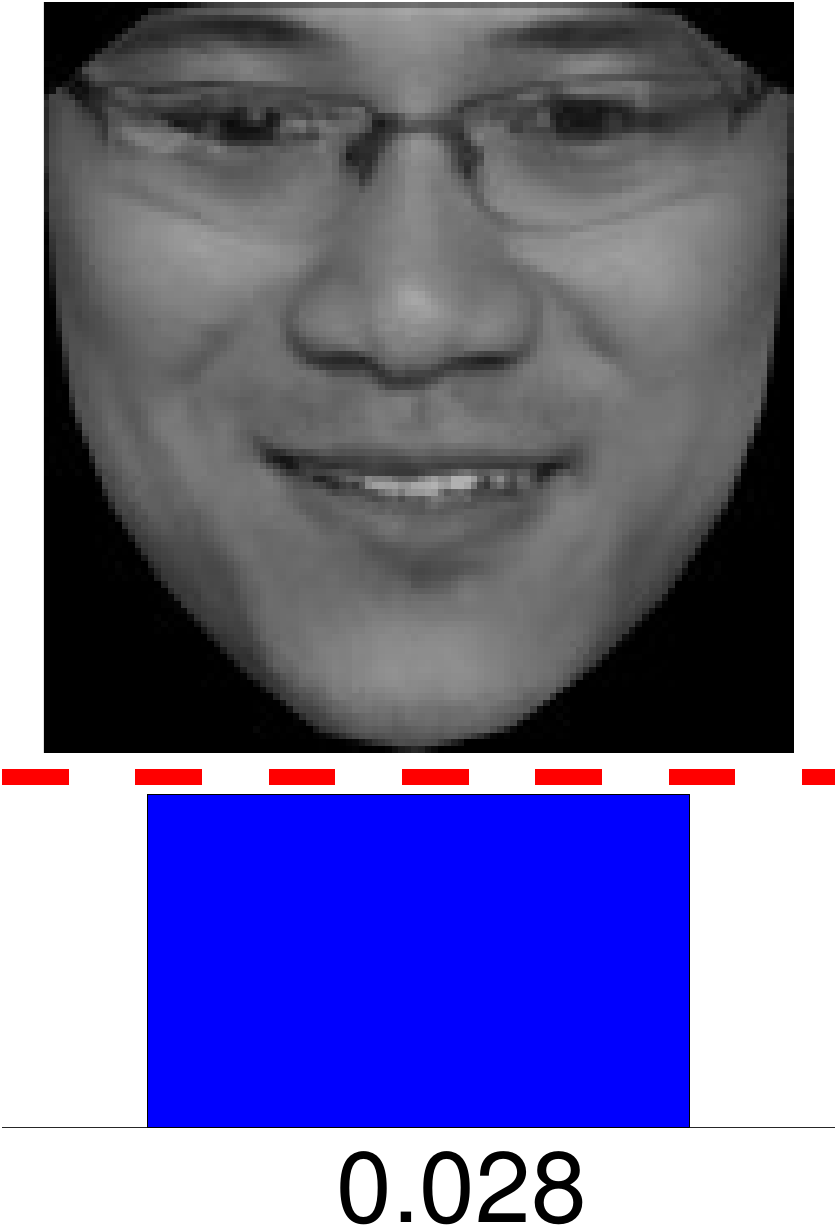}      
   \includegraphics[width=0.1\linewidth]{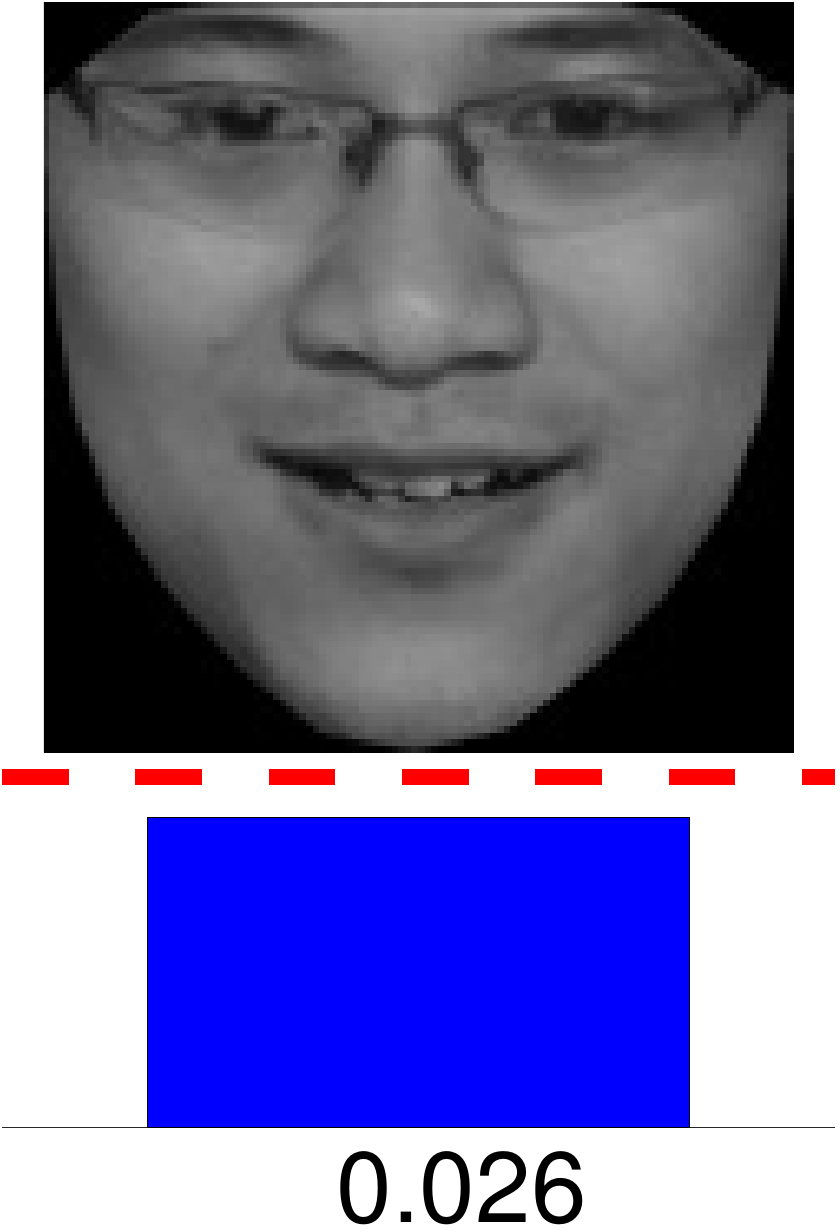}      
   \includegraphics[width=0.1\linewidth]{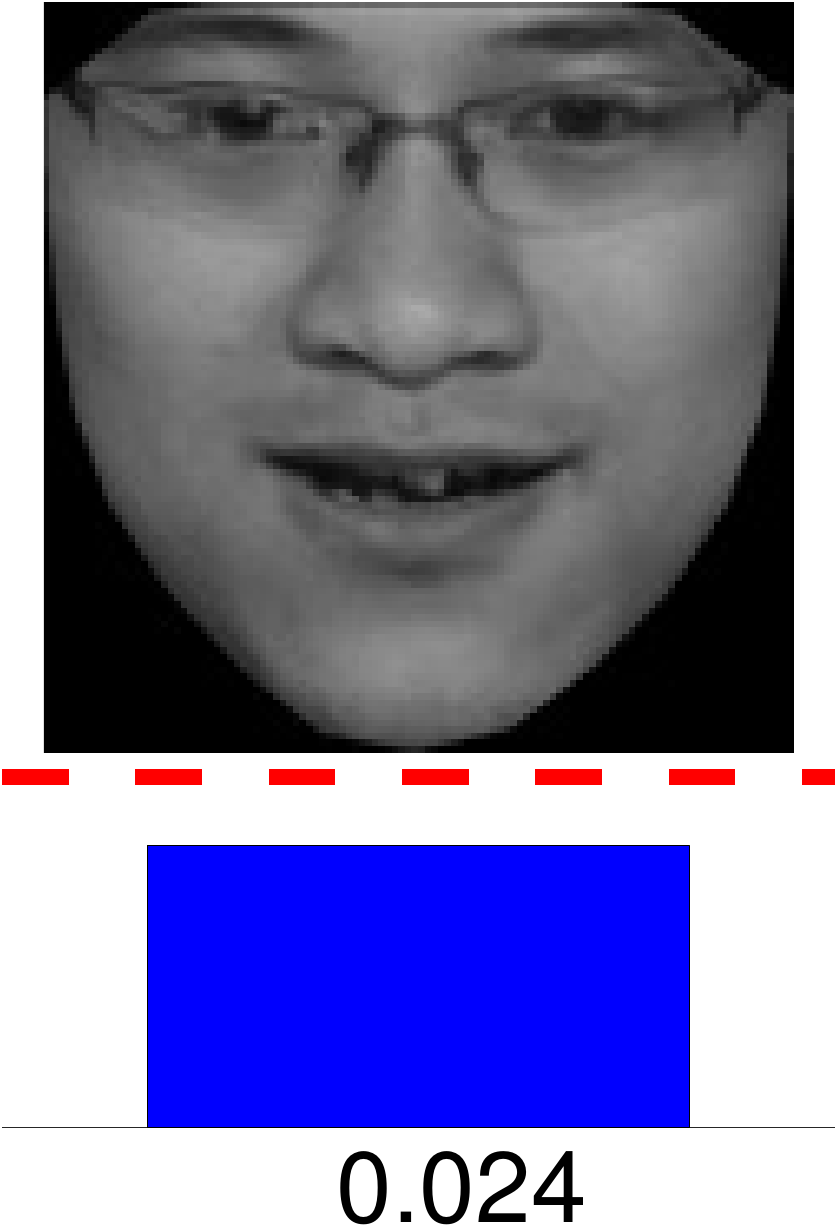} \\
   \\
   \includegraphics[width=0.1\linewidth]{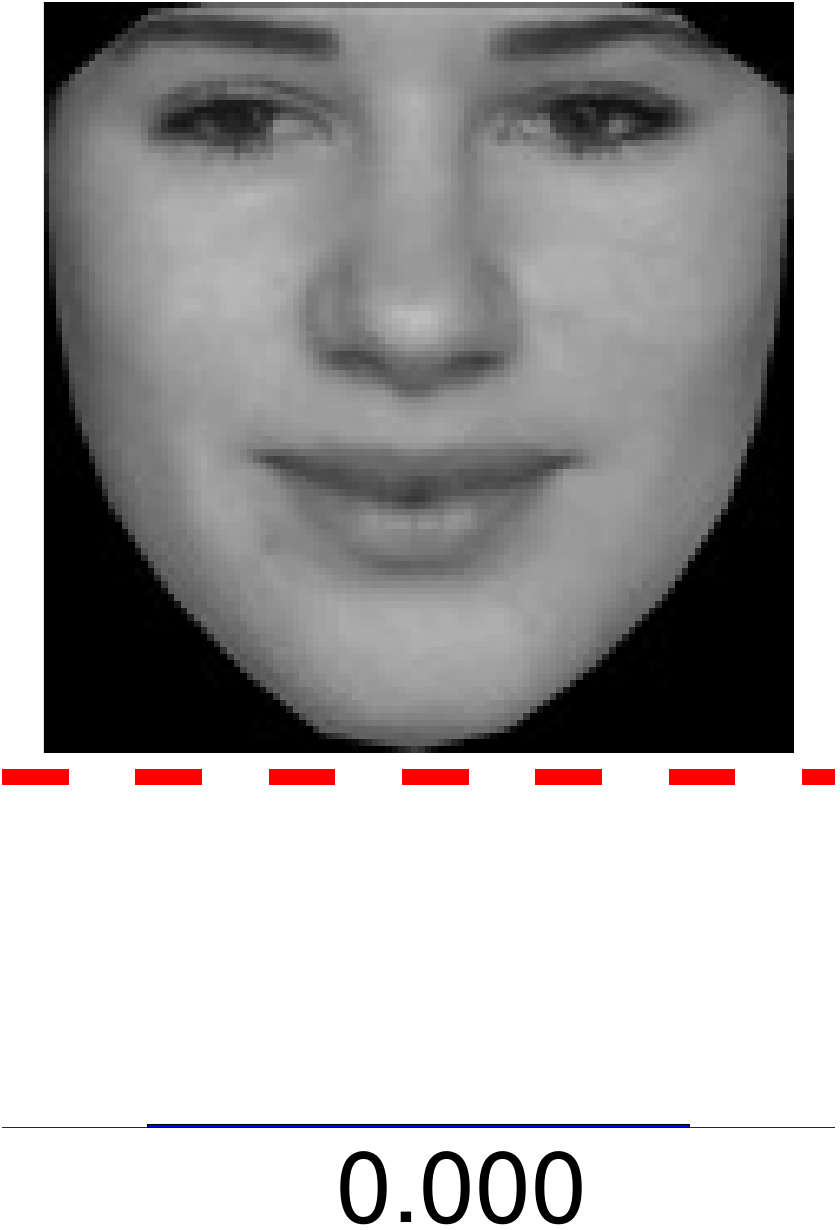} 
   \includegraphics[width=0.1\linewidth]{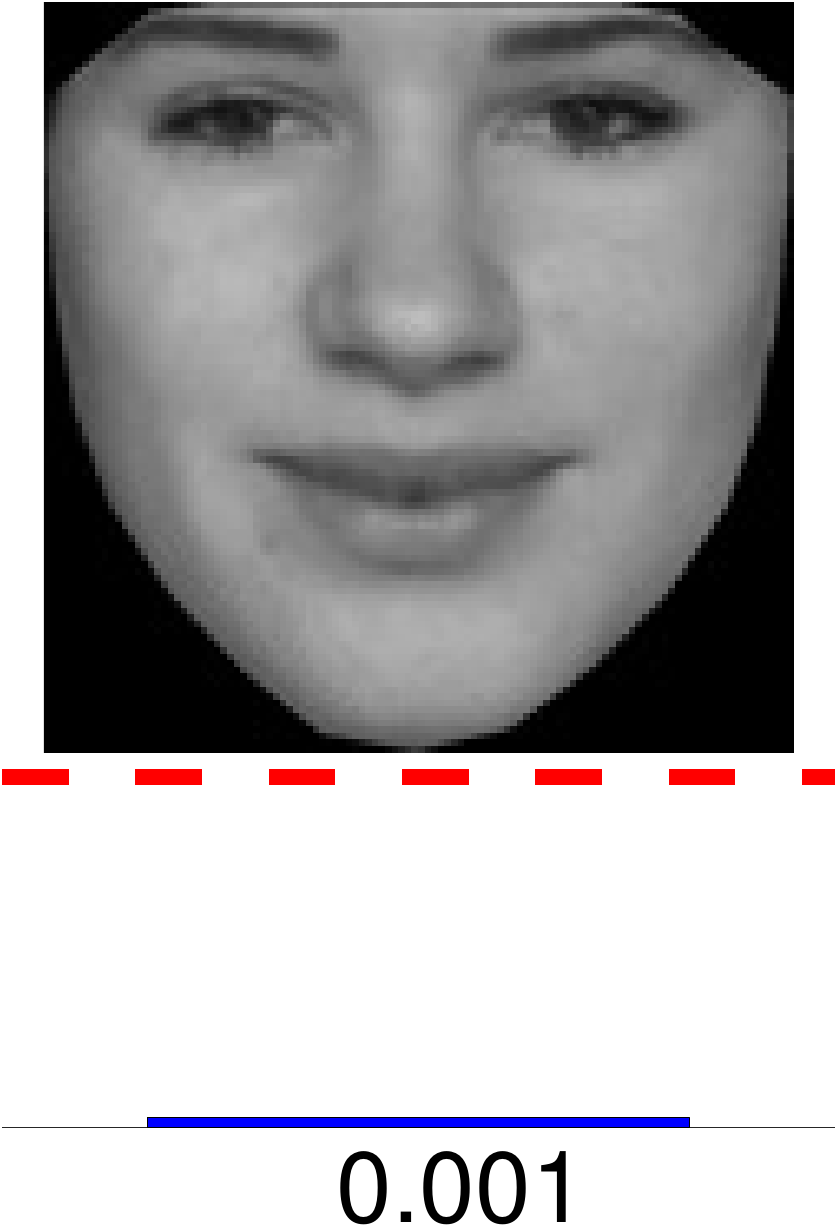}   
   \includegraphics[width=0.1\linewidth]{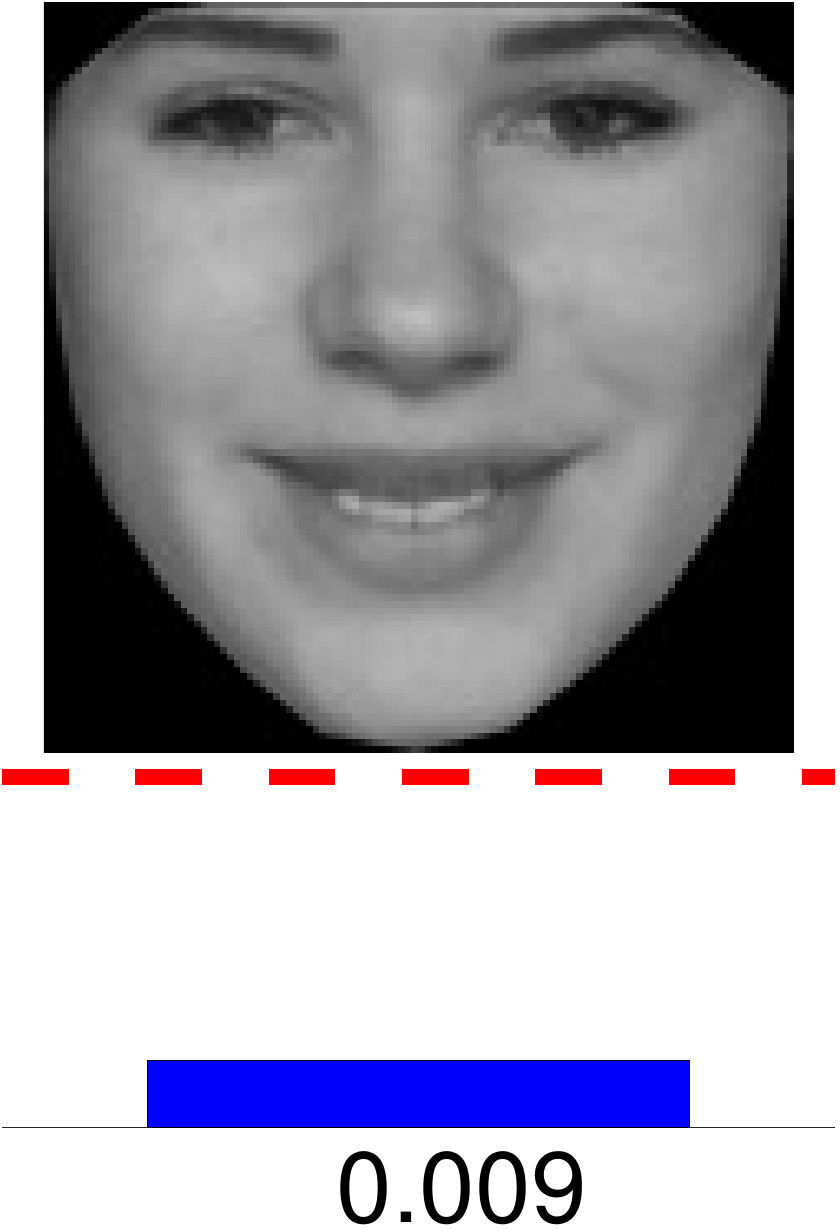}   
   \includegraphics[width=0.1\linewidth]{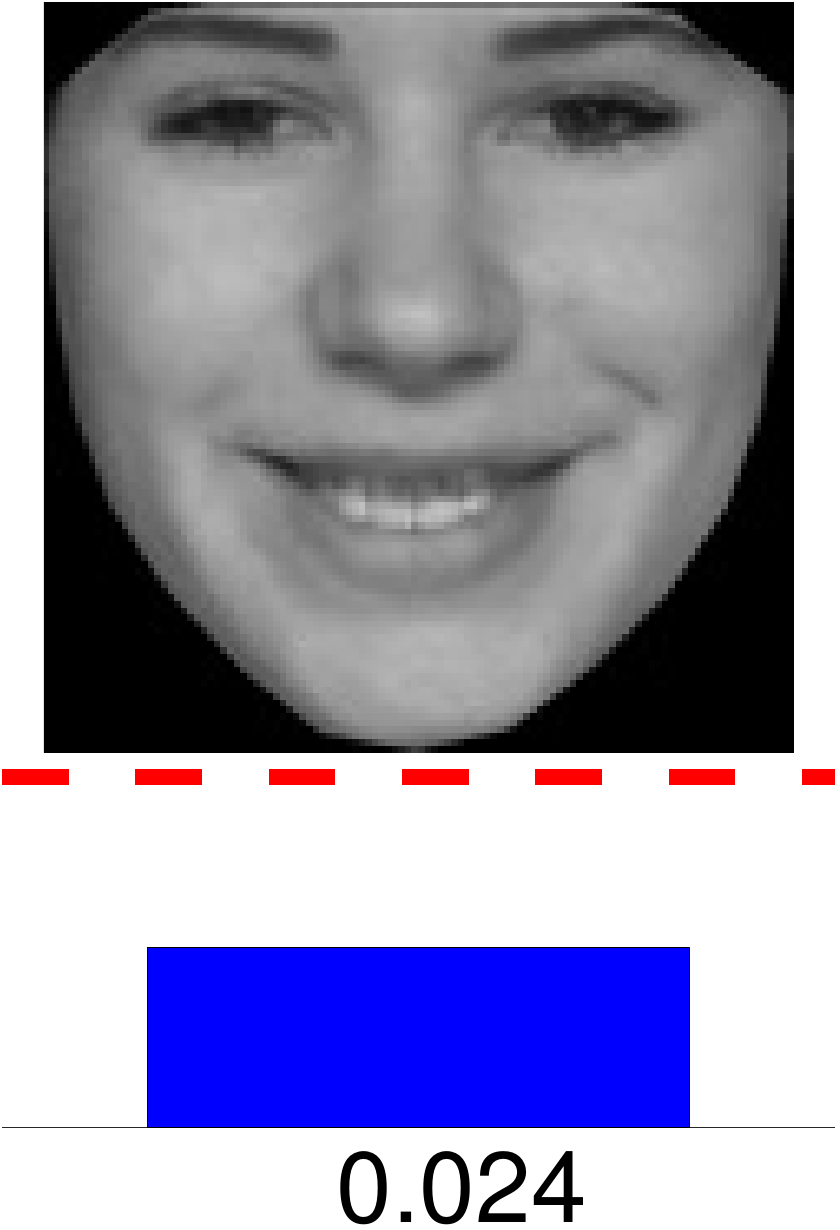}   
   \includegraphics[width=0.1\linewidth]{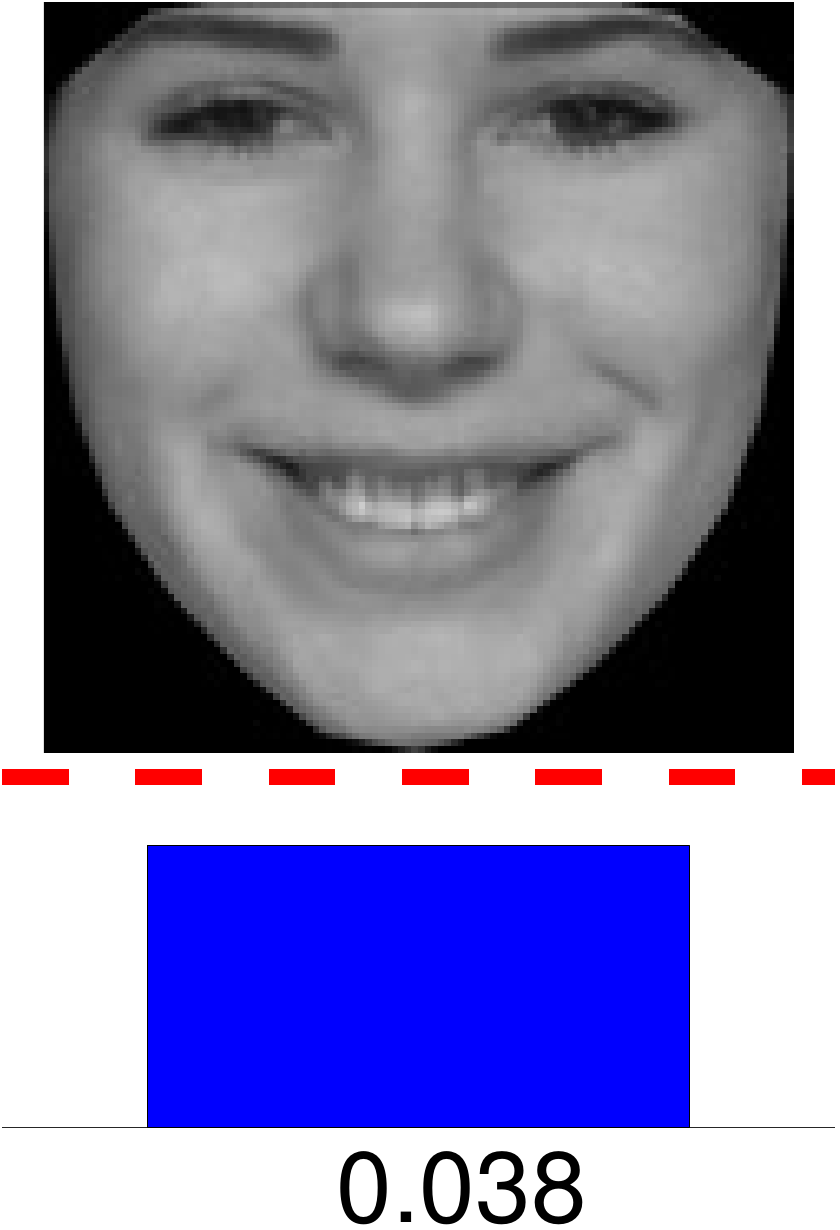}    
   \includegraphics[width=0.1\linewidth]{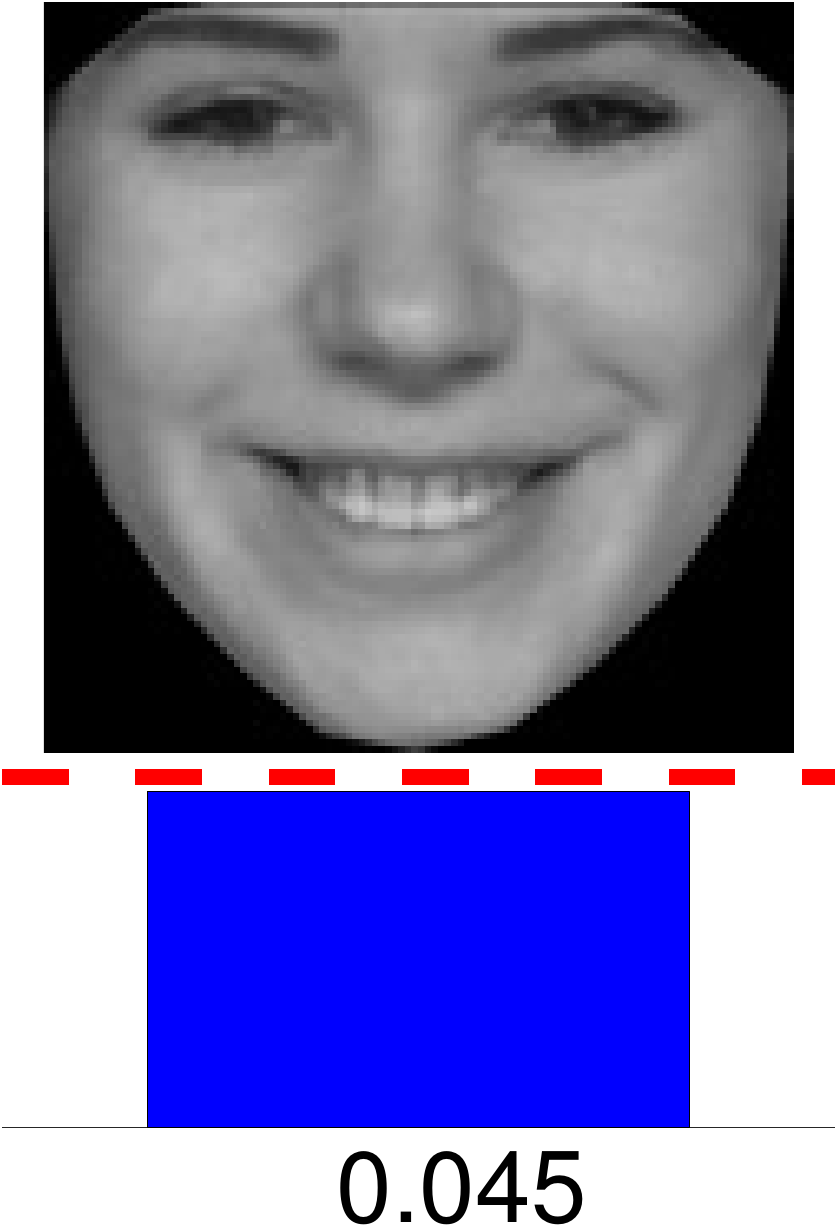}  
   \includegraphics[width=0.1\linewidth]{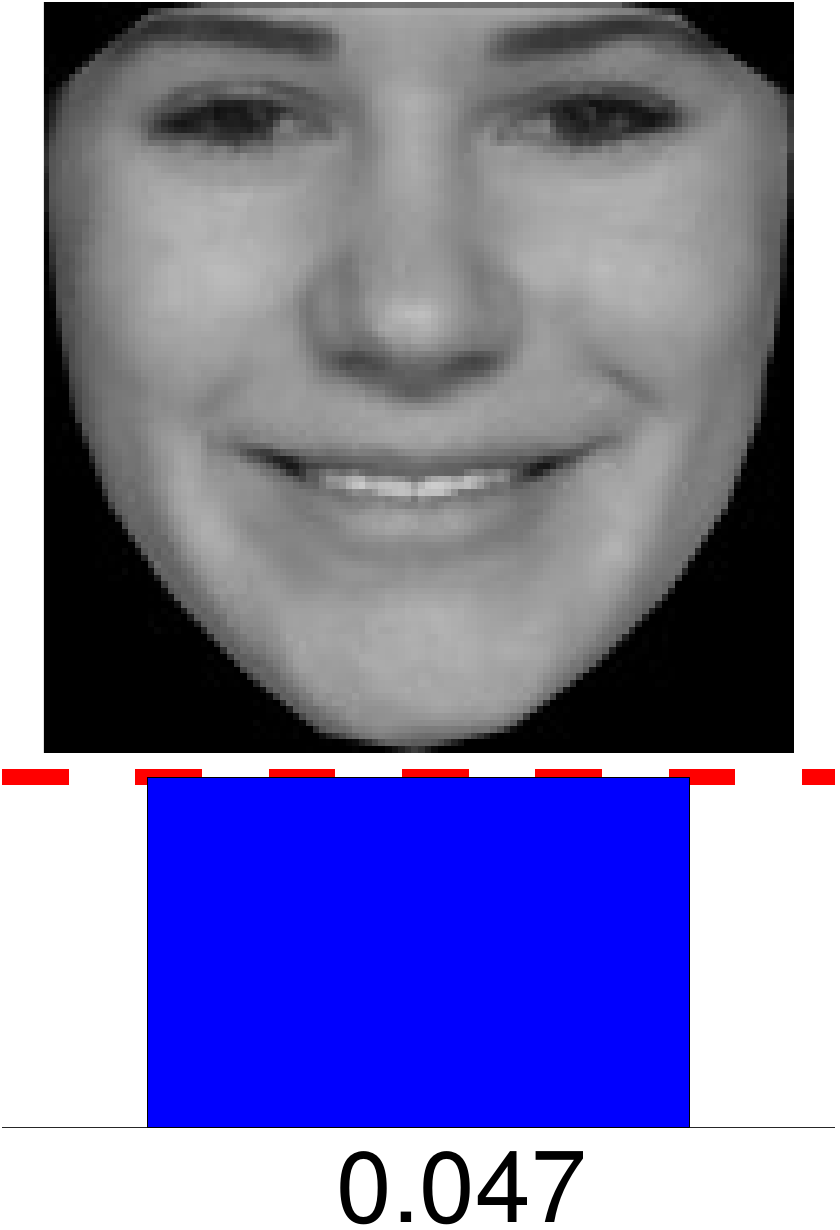}    
   \includegraphics[width=0.1\linewidth]{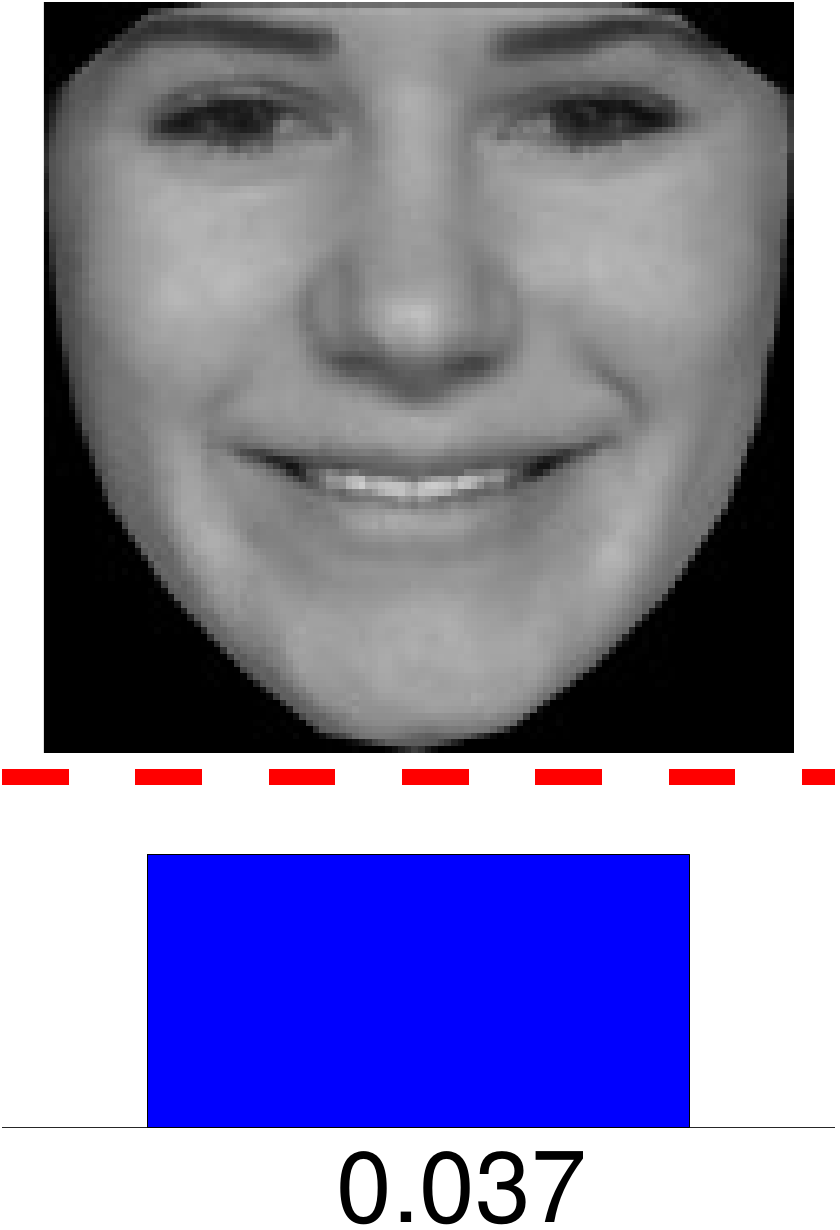}      
   \includegraphics[width=0.1\linewidth]{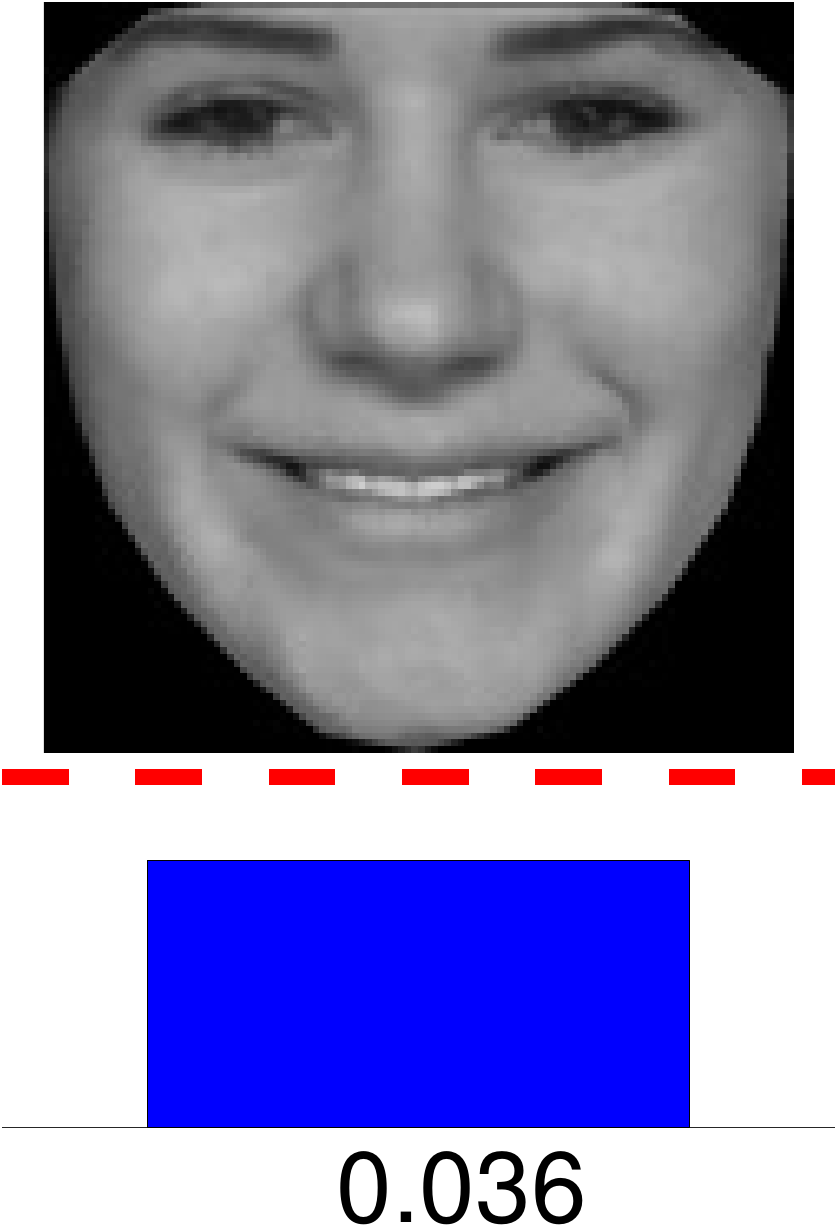}      
   \includegraphics[width=0.1\linewidth]{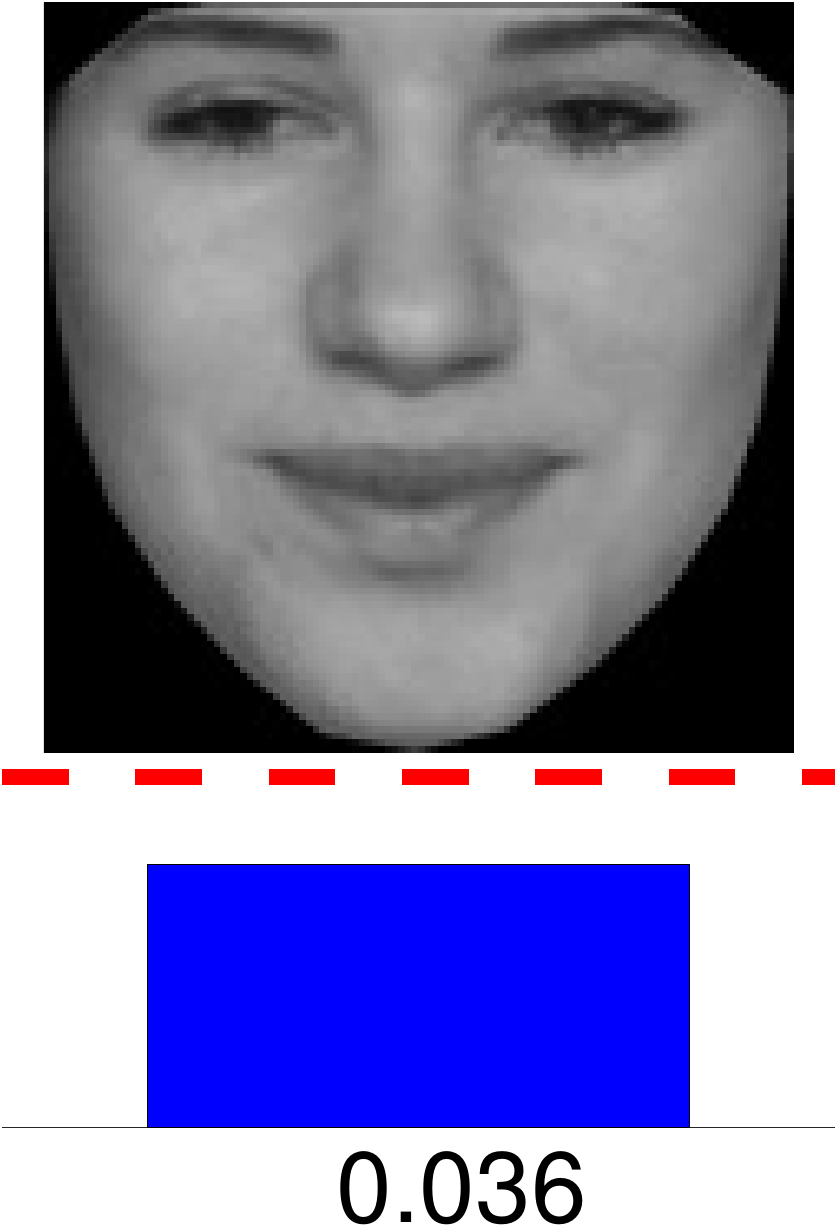} \\  
\end{tabular}}
\end{center}
\vspace{-2mm}
   \caption{The visualization of learned temporal attention weights for examples from test set of UvA-NEMO smile database. Higher bar indicates larger attention weight. The attention weight is indicated for representative frames. Our temporal attention modules is able to cut off neutral faces in the beginning phase of the smiling and assign higher value to frames with higher degree of smiling. Note that attention weights for all frames in a video are normalized (Equation~\ref{eqn:temp_att_normalize}) to make them sum up to 1. The red dash line indicates the maximum value of attention weights.}
\label{fig:temporal_att}
\end{figure*}

In Figure~\ref{fig:spatial_att_vis}, three sample images are presented to visualize the results of spatial attention module. For this module the optimal configuration is used: it uses the spatially-indexed mechanism and includes the spatial attention module after the second convolutional layer. The images on the left are the original inputs, the images in the middle are heat maps of the attention values, and the images on the right are upscaled heat maps to the original resolution.
These heat images show that our spatial attention module is able to not only discriminate the core facial region from the background accurately, but also capture the salient parts inside the facial region. Specifically, the area under eyes, nasal bridge, two nasolabial folds, and area around mouth (especially mentolabial sulcus) are detected as the salient parts. It is reasonable since these areas tend to generate wrinkles easily when smiling, which are discriminative features for the task of age estimation.

\subsubsection{The Temporal Attention Module}
\label{sec:temporal_att}
To demonstrate the temporal salience detection, 
the learned temporal attention weights for several representative frames from two video samples are shown in Figure~\ref{fig:temporal_att}. Each row 
shows the smile progression from a neutral state to smiling and back to neutral. For the neutral faces at the beginning, our module predicts very small temporal attention weights. As the degree of smiling increases, the attention weight goes up accordingly, until to the peak value. It should be noted that the attention value grows rapidly with the appearance of the two nasolabial folds, which is consistent with the facial salience captured by the spatial attention module (shown in Figure~\ref{fig:temporal_att}). Then the attention value decreases with the recession of smiling. However, the value still retains a relatively high value at the last frame. It is partially because the hidden representation of the last frames contains the information of all previous frames as well as key frames about smiling, hence they are still helpful for age estimation. Besides, the smile videos in the given database do not end with a perfectly netural face. Otherwise, the attention weight would continuously decrease for the latter neutral faces. 

\subsection{\highlight{Application to Disgust Expression}}
\highlight{To evaluate whether the proposed approach can be generalized to other facial expressions, we conduct experiments on the UvA-NEMO Disgust Database, which is composed of posed disgust expressions. Please note that the UvA-NEMO Smile and Disgust databases are the only facial expression video databases for age estimation.} 

\highlight{Table~\ref{table:disgust} presents the mean absolute errors (MAE) by our model as well as the state-of-the-art methods that handcraft several facial dynamics features and combine them with IEF~\cite{IEF} and LBP~\cite{Ojala_LBP} appearance features. Unexpectedly, our model performs worse than the state-of-the-art methods, which is in contrast to the experimental results on the UvA-NEMO smile database shown in Table~\ref{table:inter_compare2}. We consider two underlying reasons leading to poor performance of our model on disgust database. First, it is due to the relatively small data size of disgust database. The disgust database consists of 518 videos, which is only around half size of the smile database but shares the same age interval. It is revealed in Section~\ref{sec:train_size} that the training size per age significantly affects the performance of our model, which in turn explains that the poor performance of our model on the disgust database may be caused by the limited training data size. Besides, it is important to note that the carefully designed handcrafted features used in the competitor methods do not require large-scale training data since they rely on priori knowledge from the literature. Thus, we may claim that it is not fair to directly compare handcrafted features with deep models using relatively small training data.}

\highlight{Secondly, the disgust database has only posed expressions while the smile database consists of both spontaneous and posed expressions. One may speculate that the posed expressions are not as discriminative as the spontaneous ones for age estimation. However, \cite{DibekliogluICM2012} reports a better age estimation performance for posed expressions. Therefore, to further assess the negative effect of small data size by discarding the probable influence of expression spontaneity, we split the UvA-NEMO Smile Database into the spontaneous and posed smile sets, and evaluate our model solely using the posed set (643 posed smiles). Table~\ref{table:disgust} shows that our model performs much worse on the posed smiles (MAE $=6.41$ years) in comparison to the whole (spontaneous + posed) smile dataset (MAE $=4.74$ years). This result is consistent with our conjecture. Hence, we can claim that the performance our model achieves on the UvA-NEMO Disgust Database is reasonable due to the small data size, and a significant accuracy improvement can be achieved using a larger training set.}

\begin{table}[!tb]
\caption{\highlight{Mean absolute error (years) for different methods on the UvA-NEMO disgust database and the posed UvA-NEMO smile database.}}
\vspace{-.1in}
\centering
\vspace{2mm}
\renewcommand\arraystretch{1.4}
\begin{tabular}{l|l|c}
\Xhline{1.0pt}
 \highlight{\textbf{Database}} &\highlight{\textbf{Method}} & \highlight{\textbf{MAE (years)}}\\
 \hline
 \multirow{3}{*}{\tabincell{l}{\highlight{Posed Disgust}}} & $\highlight{\text{IEF+Dynamics}}$\highlight{~\cite{Hamdi2015}} & $\highlight{5.06}$ \highlight{($\pm 4.45$)}\\
 & $\highlight{\text{LBP+Dynamics}}$\highlight{~\cite{Hamdi2015}}& $\highlight{5.19}$	\highlight{($\pm 4.51$)}\\
 & \highlight{\text{SIAM (Our model)}}& $\highlight{6.96}$ $\highlight{(\pm 7.24)}$\\
\hline 
\multirow{3}{*}{\tabincell{l}{\highlight{Posed Smile}}} 
& $\highlight{\text{IEF+Dynamics}}$\highlight{~\cite{Hamdi2015}} & $\highlight{4.91}$ \highlight{($\pm 4.17$)}\\
 & $\highlight{\text{LBP+Dynamics}}$\highlight{~\cite{Hamdi2015}}& $\highlight{5.16}$	\highlight{($\pm 4.30$)}\\
 & \highlight{\text{SIAM (Our model)}}& $\highlight{6.41}$ $\highlight{(\pm 6.65)}$\\
\Xhline{1pt}
\end{tabular}
\label{table:disgust}
\end{table}

\section{Conclusion}

In this work, we present an attended end-to-end model for age estimation from facial expression videos. The model employs convolutional networks to learn the effective appearance features and feed them into recurrent networks to learn the temporal dynamics. Furthermore, both a spatial attention mechanism and a temporal attention mechanism are added to the model. The spatial attention can be integrated seamlessly into the convolutional layers to capture the salient facial regions in each single image, while the temporal attention is incorporated in recurrent networks to capture the salient temporal frames. The whole model can be trained readily in an end-to-end manner. Provided that a sufficient number of samples are available for training, we show the strong performance of our model on a large smile database. Specifically, our model makes a substantial improvement over the state-of-the-art methods. \highlight{Furthermore, we assess the applicability of the proposed method for disgust expression.}

In future work, we aim to leverage the pre-trained convolutional neural networks on large image data for the appearance learning instead of training our convolutional appearance module from scratch. This would not only accelerate the training speed but also allows employing quite deeper architectures and abundant existing image data to improve the performance of the appearance learning.

\ifCLASSOPTIONcaptionsoff
  \newpage
\fi



\bibliographystyle{IEEEtran}
\bibliography{egbib}
%



%

\begin{IEEEbiography}[{\includegraphics[width=1in,height=1.25in,clip,keepaspectratio]{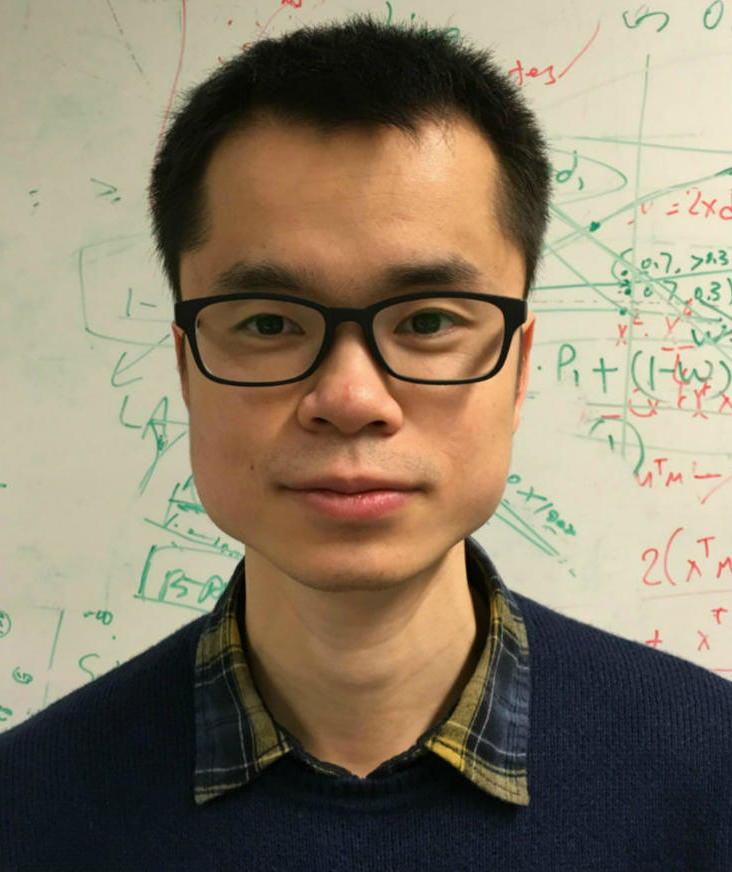}}]{Wenjie Pei} is currently an Assistant Professor with the Harbin Institute of Technology, Shenzhen, China. He received the Ph.D. degree from the Delft University of Technology, the Netherlands in 2018. Before joining Harbin Institute of Technology, he was a Senior Researcher on Computer Vision at Tencent Youtu X-Lab. In 2016, he was a visiting scholar with the Carnegie Mellon University. His research interests lie in Computer Vision and Pattern Recognition including sequence modeling, deep learning, video/image captioning, etc.
\end{IEEEbiography}
\vspace{-6mm}
\begin{IEEEbiography}[{\includegraphics[width=1in,height=1.25in,clip,keepaspectratio]{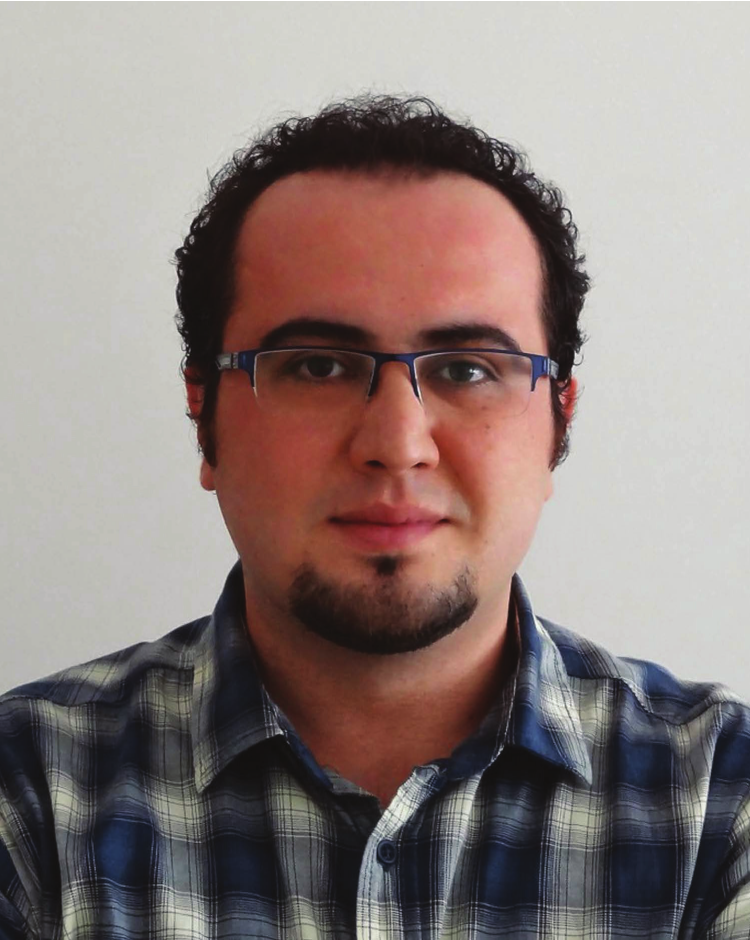}}]{Hamdi Dibeklio{\u g}lu} (S'08--M'15) is an Assistant Professor in the Computer Engineering Department of Bilkent University, Ankara, Turkey, as well as being a Research Affiliate with the Pattern Recognition \& Bioinformatics Group of Delft University of Technology, Delft, the Netherlands. He received the Ph.D. degree from the University of Amsterdam, Amsterdam, the Netherlands, in 2014. Before joining Bilkent University, he was a Postdoctoral Researcher at Delft University of Technology. His research focuses on Computer Vision, Pattern Recognition, Affective Computing, and Computer Analysis of Human Behavior. Dr. Dibeklio{\u g}lu is a Program Committee Member for several top tier conferences in these areas. He was a Co-chair for the Netherlands Conference on Computer Vision 2015, a Local Arrangements Co-chair for the European Conference on Computer Vision 2016, a Publication Co-chair for the European Conference on Computer Vision 2018, and a Co-chair for the eNTERFACE Workshop on Multimodal Interfaces 2019.
\end{IEEEbiography}
\vspace{-6mm}
\begin{IEEEbiography}[{\includegraphics[width=1in,height=1.25in,clip,keepaspectratio]{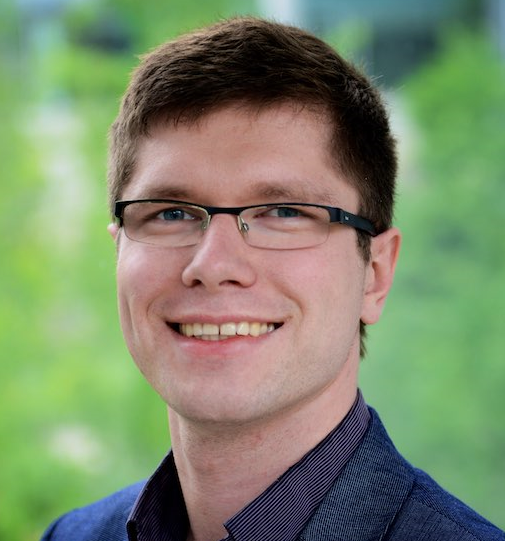}}]{Tadas~Baltru\v{s}aitis}  received the bachelor’s and PhD degrees in computer science. He is a senior scientist in the Microsoft Corporation. His primary research interests include the automatic understanding of non-verbal human behaviour, computer vision, and multimodal machine learning. Before joining Microsoft, he was a post-doctoral associate with the Carnegie Mellon University. His PhD research focused on automatic facial expression analysis in especially difficult real-world settings.
\end{IEEEbiography}
\vspace{-6mm}
\begin{IEEEbiography}[{\includegraphics[width=1in,height=1.25in,clip,keepaspectratio]{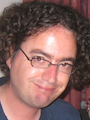}}]{David M.J. Tax} studied physics at the University of Nijmegen, The Netherlands in 1996, and received Master degree with the thesis ''Learning of structure by Many-take-all Neural Networks''. After that he had his PhD at the Delft University of Technology in the Pattern Recognition group, under the supervision of R.P.W. Duin. In 2001 he promoted with the thesis 'One-class classification'.  After working for two years as a Marie Curie Fellow in the Intelligent Data Analysis group
in Berlin, at present he is assistant professor in the Pattern
Recognition and Bioinformatics group at the Delft university of
Technology.  His main research interest is in the learning and
development of detection algorithms and (one-class) classifiers that
optimize alternative performance criteria like ordering criteria using the Area under the ROC curve or a Precision-Recall graph. Furthermore, he is lecturing in courses Pattern Recognition, Machine Learning, Probability and Statistics, and Stochastic Processes.
\end{IEEEbiography}
\vfill

\end{document}